%% file: main.tex
\RequirePackage{iftex}
\ifPDFTeX\else
  \AddToHook{package/microtype/after}{\renewcommand{\DisableLigatures}[2][]{}}
\fi
\documentclass[]{fairmeta}

\renewcommand{\sfdefault}{cmss}

\normalfont
\input{math_commands.tex}

\usepackage{amssymb}
\usepackage{textcomp}
\usepackage{tabularx}
\usepackage{colortbl}
\usepackage{longtable}
\usepackage{url}
\input{citation_links.tex}

\preto\section{\FloatBarrier}
\renewcommand{\paragraph}[1]{\vspace{1.25mm}\noindent\textbf{#1}}

\newcommand{\VLADelta}{VLA-$\Delta$}

\title{VLA-$\Delta$: Learning Physical State Transition for Vision-Language-Action Models}
\input{sections/frontmatter}
\abstract{\input{sections/abstract}}

\begin{document}
\thispagestyle{firstheader}
\maketitle
\pagestyle{empty}

\input{sections/introduction}
\input{sections/method}
\input{sections/experiments}
\input{sections/conclusion}

\subsection*{AI use statement}
Generative AI tools were used for language editing and polishing, translation, LaTeX formatting,
compilation troubleshooting, and the retrieval and discovery of related work. AI tools also
supported data annotation and quality auditing, including semantic annotation generation,
review of ambiguous or conflicting annotations against visual evidence, and bounding-box
reference generation for selected audit fields, as detailed in
Appendix~\ref{sec:appendix-annotation-pipeline}. The authors reviewed all AI-assisted work and
take responsibility for the accuracy and integrity of the final manuscript, references, data,
results, and claims.

\bibliographystyle{assets/acl_natbib}
\bibliography{references}

\clearpage
\input{sections/appendix}

\end{document}

%% file: math_commands.tex
\usepackage{amsmath,amsfonts,bm}

\def\eqref#1{equation~\ref{#1}}

\def\1{\bm{1}}

\DeclareMathAlphabet{\mathsfit}{\encodingdefault}{\sfdefault}{m}{sl}
\SetMathAlphabet{\mathsfit}{bold}{\encodingdefault}{\sfdefault}{bx}{n}



%% file: citation_links.tex
\makeatletter
\ExplSyntaxOn
\tl_new:N \l_paper_citation_name_tl
\tl_new:N \l_paper_citation_word_tl
\seq_new:N \l_paper_citation_words_seq
\cs_new_protected:Npn \paper_citation_unwrap:n #1
  { \tl_set:Nn \l_paper_citation_name_tl {#1} }
\cs_set_protected:Npn \NAT@nmfmt #1
  {
    \group_begin:
    \protected@edef\l_paper_citation_name_tl{#1}
    \exp_args:NV \tl_if_single:nT \l_paper_citation_name_tl
      {
        \exp_args:NV \tl_if_head_is_group:nT \l_paper_citation_name_tl
          { \exp_after:wN \paper_citation_unwrap:n \l_paper_citation_name_tl }
      }
    \seq_set_split:NnV \l_paper_citation_words_seq { ~ } \l_paper_citation_name_tl
    \seq_pop_left:NN \l_paper_citation_words_seq \l_paper_citation_word_tl
    \mbox{\NAT@up{{\l_paper_citation_word_tl}}}
    \seq_map_inline:Nn \l_paper_citation_words_seq
      {
        \hyper@natlinkbreak{\space}{\@citeb\@extra@b@citeb}
        \mbox{##1}
      }
    \group_end:
  }
\ExplSyntaxOff
\makeatother

%% file: sections/frontmatter.tex
\setlabdisplayname{OmniAI Group of ZJU ACES Lab}
\setuniversityname{Zhejiang University}

\author[1]{Qi Lu}
\author[1]{Jiarun Li}
\author[2]{Ce Zhang}
\author[1]{Miao Pan}
\author[1]{Yi Pan}
\author[1]{Xu Wang}
\author[1]{Wenqi Zhang}
\author[1]{Xuhong Zhang}

\affiliation[1]{Zhejiang University (\href{https://zju.edu.cn}{zju.edu.cn})}
\affiliation[2]{Central South University (\href{https://csu.edu.cn}{csu.edu.cn})}
\correspondence{\href{mailto:lu_qi@zju.edu.cn,zhangwenqi@zju.edu.cn,zhangxuhong@zju.edu.cn}{%
  \texttt{\char123 lu\_qi,zhangwenqi,zhangxuhong\char125 @zju.edu.cn}}}
\date{September 2026}

%% file: sections/abstract.tex
Vision-language-action (VLA) models transfer semantic knowledge acquired through vision-language
pretraining to robot control, achieving strong performance across various benchmarks. However,
most VLAs directly map observations to demonstrated actions without explicit supervision of
intermediate physical state changes, making them prone to overfitting observation--action
correspondences. Some methods introduce intermediate reasoning or future-state prediction in
language, visual, or latent spaces to guide action generation, but do not explicitly supervise the
physical state transition itself. We therefore propose \VLADelta{}, a general framework that uses
physical changes between current and future states as explicit learning targets. To support this
objective, we construct a large-scale pretraining dataset and derive physical state-transition
annotations covering task-relevant semantics and visual motion changes, providing more direct
supervision for continuous action generation. During inference, the model generates actions
directly from current observations without additional inference overhead. \VLADelta{} achieves
performance competitive with the state of the art on LIBERO and RoboTwin, and outperforms
$\pi_{0.5}$ in real-robot experiments. Our code is available
at \url{https://github.com/ZJU-OmniAI/VLA-Delta}.

%% file: sections/introduction.tex
\section{Introduction}

Building generalist robot policies that follow natural-language instructions in diverse environments
is a long-standing goal of embodied intelligence. Vision-language-action (VLA)
models~\citep{RT2023,kim2024openvla,GR00T2025,Pi02024,intelligence2025pi_,zheng2026x,community2026starvla,li2024cogact} have recently emerged as a promising paradigm toward
this goal. Built on pretrained vision-language models (VLMs), VLAs inherit rich semantic knowledge
and visual understanding from large-scale vision-language pretraining. They use the VLM backbone
to encode task instructions and visual observations, mapping these representations to robot actions
and achieving strong performance across diverse benchmarks.

Despite this progress, VLAs often exhibit failures revealing a limited understanding of physical
dynamics. For example, a policy may close its gripper before reaching the target object, or continue
toward the placement location after a failed grasp. In such cases, the policy appears to
replay a memorized trajectory rather than respond to the evolving scene. These failures suggest
that reliable manipulation requires reasoning about \textbf{how the robot and its environment should
change} as the task progresses. A key reason is that most VLAs directly map observations to
demonstrated actions and \textbf{lack explicit supervision of the intermediate physical state
changes}. They tend to overfit observation--action correspondences instead of capturing
how physical states change over time.

\begin{figure}[t]
    \centering
    \includegraphics[width=\linewidth]{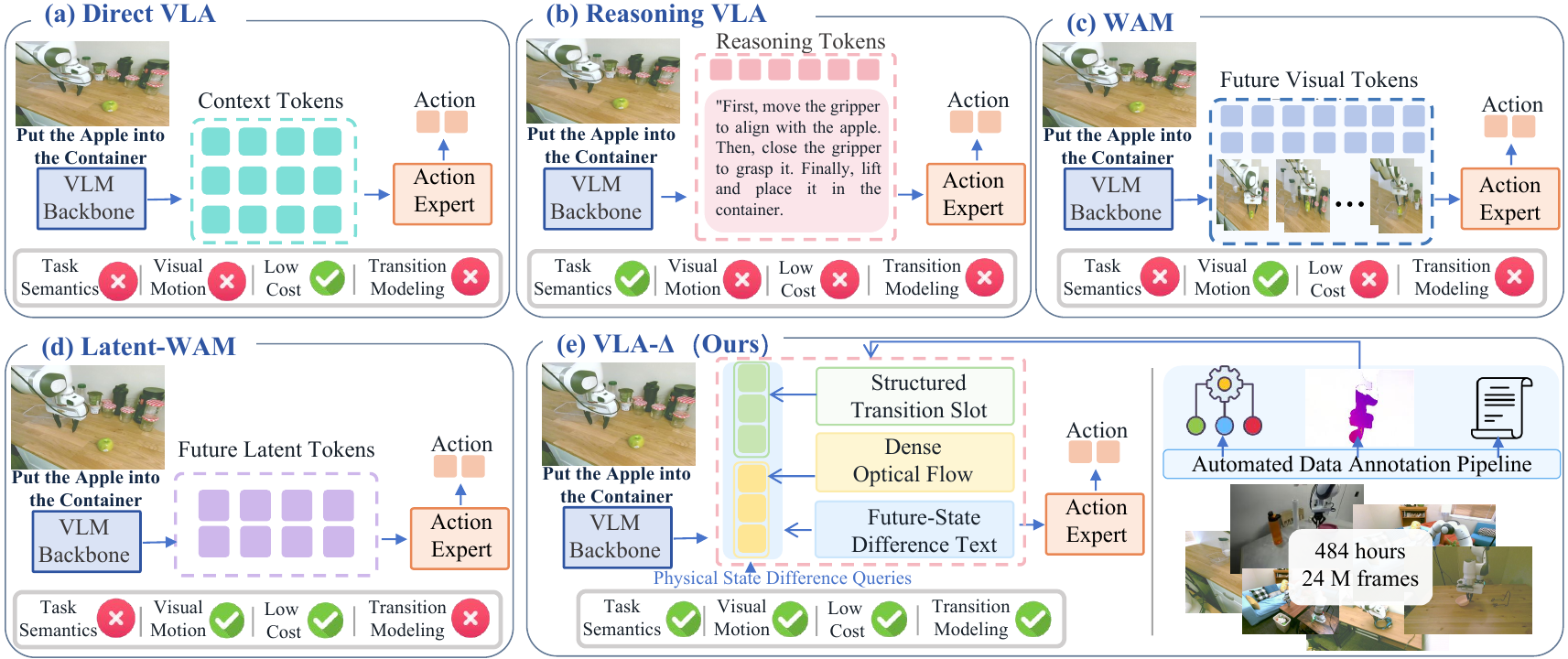}
    \caption{Comparison of transition modeling paradigms in vision--language--action policies.
    (a) Direct VLA maps observations and instructions directly to actions. (b) Reasoning VLA
    describes task plans and subgoals in language space. (c) WAM predicts future visual states,
    while (d) Latent-WAM predicts future latent representations. (e) VLA-$\Delta$ models physical
    state changes through multilevel supervision.}
    \label{fig:introduction}
\end{figure}

As illustrated in Fig.~\ref{fig:introduction}, existing studies
\citep{Li2026ZR0,WorldVLA2025,VLAJEPA2026,JEPAWAM2026} introduce intermediate reasoning or future-state
prediction to guide action generation by implicitly modeling state changes. (b) Reasoning VLAs such as
ECoT~\citep{ECoT2024} autoregressively infer task plans and object locations before generating actions. However,
their chain-of-thought supervision focuses on high-level task planning and state reasoning, together
with sparse visual localization. It describes what should be done next, but not how the physical
states of the robot and objects change between current and future states. (c) World Action
Models (WAMs)~\citep{cai2026internvla} model state changes by generating future images. These methods provide denser visual
details, but future-image reconstruction mainly enforces consistency in scene appearance and does
not explicitly identify which changes are necessary for task completion. Moreover, generating
future images typically introduces additional inference cost. (d) Latent-WAMs~\citep{chen2026lawam} move future prediction
into a latent space, reducing the cost of visual reconstruction. However, their latent
representations still model only future visual states, rather than signals tied to physical states
and motion interactions. Overall, these approaches supervise either \textcolor{cyan}{high-level task semantics} or
\textcolor{cyan}{future visual appearance}, but rarely supervise \textcolor{red}{the physical state
transition itself}, i.e., how object poses, spatial relations, and robot--object contacts change
from the current state to the future. Yet it is precisely these physical changes that directly define which
actions the robot must take.

To bridge this gap, we propose to \textbf{explicitly model physical state transitions as learning targets}.
Such transitions should be both linked to task intent and grounded in concrete visual
information: task semantics alone cannot specify how objects move, while visual appearance alone
cannot indicate which changes matter for the task. Our core design is to extract physical
state-transition information between the current and future states, encompassing both
task-relevant semantics and visual motion changes. We aim to learn transition representations that capture task semantics while preserving motion details, providing more direct supervision for continuous action generation.

We therefore introduce VLA-$\Delta$, a general framework using physical state transitions as
explicit learning targets. The framework incorporates learnable Physical State Difference
Queries into the vision-language backbone to form a unified transition representation conditioned
on current observations and task instructions. We shape this representation through two supervision
pathways: direct prediction and autoregressive generation. \textcircled{1}~\textbf{Structured Transition
Slots} provide structured supervision for variables including spatial positions and contact states,
encouraging a semantic division of labor among Queries and giving the transition representation
clear physical meaning. \textcircled{2}~\textbf{Optical Flow} provides dense visual motion supervision,
encouraging the representation to preserve motion details of objects and grippers.
\textcircled{3}~\textbf{Future-State Difference Text} provides coherent semantic supervision for task
progress, spatial displacement, contact changes, and associated physical constraints through
autoregressive generation. Together, these three signals guide the Query representations
to jointly encode task semantics and visual motion changes, supporting continuous action generation.

We construct a pretraining dataset comprising approximately 484 hours and 24 million frames and
derive physical state-transition supervision signals from it. The three supervision signals are introduced only
during pretraining, while downstream fine-tuning uses only the action-learning objective. This
training scheme allows the physical state-transition knowledge acquired during pretraining to
provide a representational foundation for downstream policy learning and adapt to the continuous
control requirements of specific tasks. During deployment, the model directly generates actions
without decoding intermediate supervision targets.

VLA-$\Delta$ achieves strong performance in both simulation and real-robot experiments.
VLA-$\Delta$ achieves a success rate of 98.40\% on LIBERO and outperforms all compared methods
under both the Clean and Random settings on RoboTwin. In real-robot experiments, it outperforms
$\pi_{0.5}$ by 5.5\%. These results suggest that physical state-transition supervision during
pretraining helps the model focus on task-relevant motion changes, enabling it to benefit from
pretrained transition representations during downstream fine-tuning.

Our main contributions are as follows:
\begin{enumerate}
    \item We propose VLA-$\Delta$, a general framework using physical state transitions as
    explicit learning targets. By explicitly modeling task semantics and visual motion
    changes, it provides more direct supervision for continuous action generation.

    \item We curate approximately 484 hours and 24 million frames of robot demonstrations and
    develop a complete automated data annotation pipeline to generate large-scale, multilevel
    physical state-transition annotations for model pretraining.

    \item VLA-$\Delta$ achieves the highest success rates among the compared methods on LIBERO
    and RoboTwin and outperforms $\pi_{0.5}$ in real-robot experiments.
\end{enumerate}

%% file: sections/method.tex
\section{Method}

\subsection{Overall Architecture of \texorpdfstring{\VLADelta{}}{VLA-Δ}}

\begin{figure}[t]
    \centering
    \includegraphics[width=\linewidth]{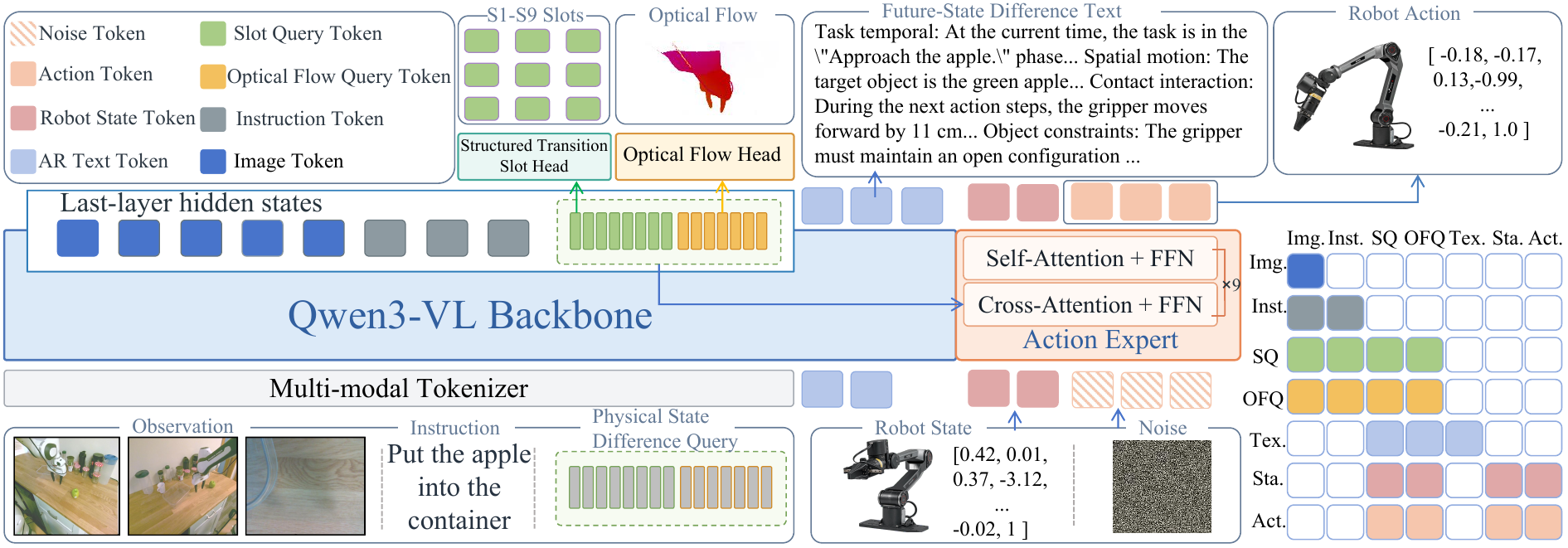}
    \caption{Overall architecture of \VLADelta{}. Given multi-view observations, a task instruction, and the robot state,
    the model jointly learns a physical state transition representation from Structured Transition Slots,
    Optical Flow, and Future-State Difference Text supervision, and uses these representations to condition continuous action generation.
    The right panel illustrates the attention-visibility constraints among the different information streams.}
    \label{fig:architecture}
\end{figure}

As illustrated in Fig.~\ref{fig:architecture}, \VLADelta{} employs learnable Physical State
Difference Queries to organize vision-language information into a physical state-transition
representation for continuous action generation. The overall architecture follows the
vision-language backbone and action-expert design of GR00T N1 \citep{GR00T2025}, using
Qwen3-VL-2B-Instruct and a 9-layer Diffusion Transformer.

At time $t$, the model receives multi-view visual observations $o_t$, a task instruction $\ell$,
and the robot proprioceptive state $r_t$, and predicts a continuous action chunk of length $H_a$,
$A_t=(a_t,\ldots,a_{t+H_a-1})$. The visual observations and task instruction are first encoded
into a contextual sequence $C_t$, which is then fed into the vision-language backbone $F_\theta$
together with the learnable Query parameters $Q\in\mathbb{R}^{N_q\times d}$. We extract the
final-layer hidden states at the Query positions to obtain the transition representation $Z_t$:

\begin{equation}
Z_t = \operatorname{Readout}_{Q}\!\left(F_\theta([C_t;Q])\right)
\in \mathbb{R}^{N_q \times d}.
\label{eq:query-readout}
\end{equation}

We shape this representation through two pathways: direct prediction and autoregressive
generation. The direct prediction pathway reads the designated Query hidden states to predict
structured variables, $\hat{y}_t=h_{\mathrm{slot}}(Z_t^{\mathrm{slot}})$, where
$\hat{y}_t=(\hat{y}_{t,1},\ldots,\hat{y}_{t,9})$, and dense optical flow,
$\hat{U}_t=h_{\mathrm{flow}}(Z_t^{\mathrm{flow}})$.

The autoregressive generation pathway predicts Future-State Difference Text,
$\hat{T}_t=\operatorname{ARDecode}_{\theta}(Q;M_{\mathrm{attn}})$. Under the attention constraints
illustrated in the right panel of Fig.~\ref{fig:architecture}, text positions can attend only to
the Queries and preceding text, and cannot directly attend to the context $C_t$. Therefore,
information from the current observations and task instruction must pass through the Query
representation before contributing to difference-text prediction.

The action expert attends to $Z_t$ through cross-attention and combines it with the independently
encoded robot proprioceptive state $E_\eta(r_t)$. It then transforms Gaussian noise into a
continuous action chunk through flow matching:
$\hat{A}_t=\operatorname{FMDecode}_{\psi}\!\left(\xi;Z_t,E_\eta(r_t)\right)$,
$\xi\sim\mathcal{N}(0,I)$.

The three auxiliary supervision signals are used jointly with action learning only during
pretraining, while downstream fine-tuning optimizes only the action-learning objective. During
deployment, the model retains the Query representation extraction and action-generation pathways
without decoding the auxiliary supervision targets.

\suppressfloats[t]
\begin{figure}[t]
    \centering
    \includegraphics[width=\linewidth]{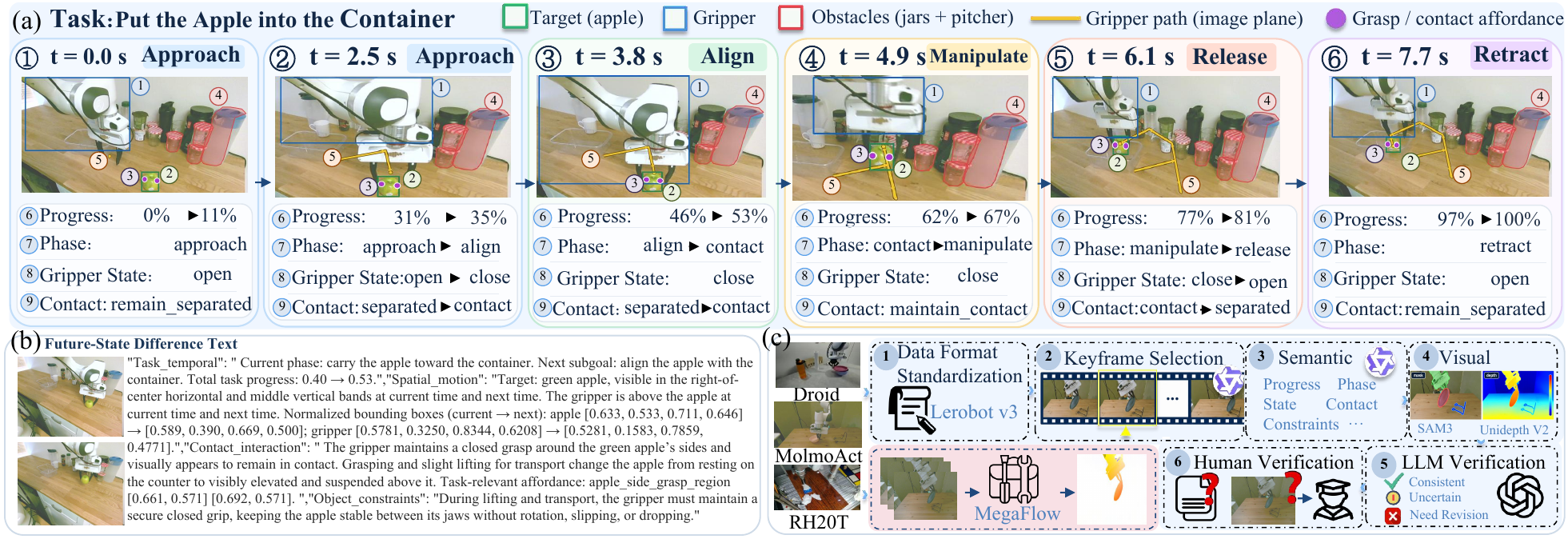}
    \captionsetup{justification=centering}
    \caption{Physical state-transition annotations and the automated data annotation pipeline.}
    \label{fig:structured-slots}
\end{figure}

\subsection{Multilevel Physical State Transition Learning}

\textbf{Structured Transition Slots Supervision.}
Physical State Difference Queries lack explicit constraints on their semantic roles when trained
solely through action prediction. To encourage transition representations to encode physically
meaningful information, we introduce Structured Transition Slots, which organize task-relevant
state variables and their changes into nine categories of supervision targets with predefined
semantic roles: task progress, task phase, target position, gripper position, motion and relative
displacement, interactable position, gripper state transitions, contact transitions, and obstacle
risk. This supervision encourages the assigned Queries to develop a soft semantic
division of labor while preserving information exchange among Queries, allowing them to jointly
support action generation. Figure~\ref{fig:structured-slots} presents an example of Structured
Transition Slot annotations for a complete manipulation episode.
Further details appear in Appendix~\ref{sec:appendix-state-transition-annotations}.

We assign a subset of Queries to these structured targets. For each slot category, a lightweight
prediction head reads the corresponding Query hidden representations to output $\hat{y}_i$.
The overall structured objective is
$\mathcal{L}_{\mathrm{slot}} = \sum_{i=1}^{9} w_i\,\mathcal{L}_i(\hat{y}_i, y_i)$,
where $w_i$ denotes the weight assigned to each target category. These structured targets provide
direct supervision for learning key state variables.

\textbf{Dense Optical Flow Supervision.}
Structured Transition Slots supervise predefined state variables and their changes, but do not
fully capture fine-grained visual motion during object movement, gripper approach, and local
alignment. To complement this information, we introduce Optical Flow supervision using dense
two-dimensional displacements between current and subsequent observations from the same viewpoint,
encouraging Query representations to preserve local motion details in the image plane. We use a
lightweight optical flow decoder that reads only the final hidden states of the last
$N_{\mathrm{flow}}$ difference queries and decodes their motion information into dense optical flow.

The total loss is defined as
$\mathcal{L}_{\mathrm{flow}} = \mathcal{L}_{\mathrm{all}}
+ \beta_{\mathrm{motion}}\mathcal{L}_{\mathrm{motion}}$,
where $\mathcal{L}_{\mathrm{all}}$ is the average error over all valid pixels, and
$\mathcal{L}_{\mathrm{motion}}$ is the average error over motion pixels whose target flow magnitude
exceeds a predefined threshold.

Optical flow alone does not distinguish task-relevant motion from other image motion. We therefore
combine it with Structured Transition Slots and Future-State Difference Text, jointly constraining
transition representations through state semantics and visual motion objectives.

\textbf{Future-State Difference Text Supervision.}
Structured Transition Slots and Optical Flow supervise physical state changes through predefined
state variables and dense visual displacements, respectively. To better support action generation,
transition representations also need to capture the relationships among these changes, including
how object motion and contact transitions jointly advance the current subgoal and which physical
constraints must be satisfied throughout this process. We therefore introduce Future-State
Difference Text, which uses coherent natural language to jointly describe task progress, spatial
changes, contact transitions, and associated constraints between current and future states.
Figure~\ref{fig:structured-slots}(b) presents an example of Future-State Difference Text.

We use the VLM's native autoregressive objective to learn Future-State Difference Text $T$.
Cross-entropy is computed only over valid target text positions $\Omega_T$:
$\mathcal{L}_{\mathrm{Text}} = -\frac{1}{|\Omega_T|}
\sum_{j\in\Omega_T} \log p_\theta\!\left(t_j \mid C,Q,t_{<j};M_{\mathrm{attn}}\right)$.
Current vision-language information is routed exclusively through Queries for difference text
prediction.

This supervision encourages Queries to capture the relationships between local state changes and
overall task progression, providing transition representations with coherent task semantics for
continuous action generation.

\textbf{Action Generation Conditioned on the Transition Representation.}
The action expert combines the transition representation $Z_t$ with the independently encoded
proprioceptive state $E_\eta(r_t)$ to predict a continuous action chunk. We adopt the Flow
Matching objective for continuous actions. Given a normalized demonstration
action chunk $A_t$, we sample Gaussian noise $\xi$ of the same shape and an interpolation time
$\tau\in[0,1]$, and construct
$A_t^\tau = (1-\tau)\xi + \tau A_t$,
$v_t^\star = A_t - \xi$, $\xi\sim\mathcal{N}(0,I)$.

The action expert $v_\psi$ predicts the vector field from the noisy action, time, and
conditional representation, optimized by
$\mathcal{L}_{\mathrm{act}} = \mathbb{E}\left[
\frac{1}{|\Omega_a|}\sum_{k\in\Omega_a}
\left(
\left[v_\psi\!\left(A_t^\tau,\tau\mid Z_t,E_\eta(r_t)\right)\right]_k
- [v_t^\star]_k
\right)^2
\right]$.

\textbf{Training and Inference.}
During joint pretraining, we optimize physical state-transition prediction and continuous action
generation using the following objective:

\begin{equation}
\mathcal{L} = \lambda_{\mathrm{Text}}\mathcal{L}_{\mathrm{Text}}
+ \lambda_{\mathrm{slot}}\mathcal{L}_{\mathrm{slot}}
+ \lambda_{\mathrm{flow}}\mathcal{L}_{\mathrm{flow}}
+ \lambda_{\mathrm{act}}\mathcal{L}_{\mathrm{act}}.
\label{eq:joint-pretraining-loss}
\end{equation}

When fine-tuning on demonstrations from downstream benchmarks, we optimize only the flow-matching
action objective, $\mathcal{L}=\mathcal{L}_{\mathrm{act}}$. This allows the transition
representations learned during pretraining to adapt to downstream tasks through action supervision,
without requiring additional transition annotations for downstream demonstrations.

During inference, the VLM extracts the transition representation $Z_t$ from the current visual
observations and task instruction. The action expert then generates continuous action chunks via
flow matching, conditioned on $Z_t$ and the robot proprioceptive state.

\subsection{Automated Data Annotation Pipeline}
\label{sec:pretraining-dataset}

As shown in Figure~\ref{fig:structured-slots}(c), we develop an automated annotation pipeline to produce standardized physical state-transition
annotations for robot demonstrations from multiple sources. First, we convert all data sources
into the unified LeRobot v3 trajectory format. For keyframe selection, we use Qwen3.8 to segment
demonstration videos into manipulation stages based on task instructions, identifying processes
such as approaching, aligning, and grasping. Stage boundaries serve as keyframe candidates, and
additional keyframes are sampled adaptively according to the complexity of the operations within
each stage. For semantic annotation, we use Qwen3.8 to generate structured semantic labels. For
visual annotation, we use the object descriptions in these labels to prompt SAM 3
\citep{carion2025sam3} to obtain bounding boxes and segmentation masks for target objects,
grippers, and obstacles. We then use
UniDepth V2 \citep{piccinelli2025unidepthv2} to estimate image depth and extract depth information
near the center of each target object. Finally, we align and integrate semantic and geometric
information from current and future time steps according to predefined rules, producing unified
physical state-transition annotations and constructing training samples from them.
For optical flow annotation, we use MegaFlow
\citep{zhang2026megaflow} to estimate dense two-dimensional displacements offline between the
current and subsequent frames from the same camera view.

Quality control is integrated throughout the pipeline. We first use automated checks to identify
formatting errors, temporal misalignment, invalid coordinates, inconsistent object identities, and
depth anomalies. GPT-5.6 then reviews complex cases or conflicting annotations against visual
evidence on a field-by-field basis and flags uncertain information. Finally, we conduct a stratified
manual audit to assess annotation quality. We describe the annotation pipeline in detail in
Appendix~\ref{sec:appendix-annotation-pipeline}.

%% file: sections/experiments.tex
\section{Experiments}

\subsection{Experimental Setup}

\textbf{Implementation and training.}
\VLADelta{} is built on the GR00T N1 architecture. The vision-language model is initialized with
the weights of ZR-0$_{\mathrm{base}}$ \citep{Li2026ZR0}. We pretrain the model with the joint
training objective on demonstrations from DROID, MolmoAct, and RH20T, annotated using the
automated annotation pipeline described in Section~\ref{sec:pretraining-dataset}.

\textbf{Simulation and real-world platforms.}
We evaluate \VLADelta{} on the LIBERO~\citep{liu2023libero} and RoboTwin 2.0~\citep{chen2025robotwin}
and assess its generalization on LIBERO-Plus~\citep{fei2025libero}. We further conduct real-world
experiments on the AgileX PiPER 6-DoF dual-arm robot platform. Task success rate serves as the
primary evaluation metric. Detailed training and evaluation settings are provided in
Appendix~\ref{sec:appendix-evaluation-protocol}.

\subsection{Simulation Benchmark Evaluation}

\input{tables/table1_libero}

\noindent
\begin{minipage}[t]{0.48\textwidth}
    \vspace{0pt}
    \paragraph{\textbf{LIBERO and LIBERO-Plus.}}
    We first evaluate the model on LIBERO. As shown in Table~\ref{tab:libero_main}, \VLADelta{} achieves an average success rate of
    \textbf{98.4\%}, the highest among all compared methods.
    To further evaluate generalization under out-of-distribution environmental changes, we test
    the model on four perturbation types in LIBERO-Plus: Language, Light, Background, and Layout.
\end{minipage}\hfill
\begin{minipage}[t]{0.48\textwidth}
    \vspace{0pt}
    \centering
    \includegraphics[width=\linewidth]{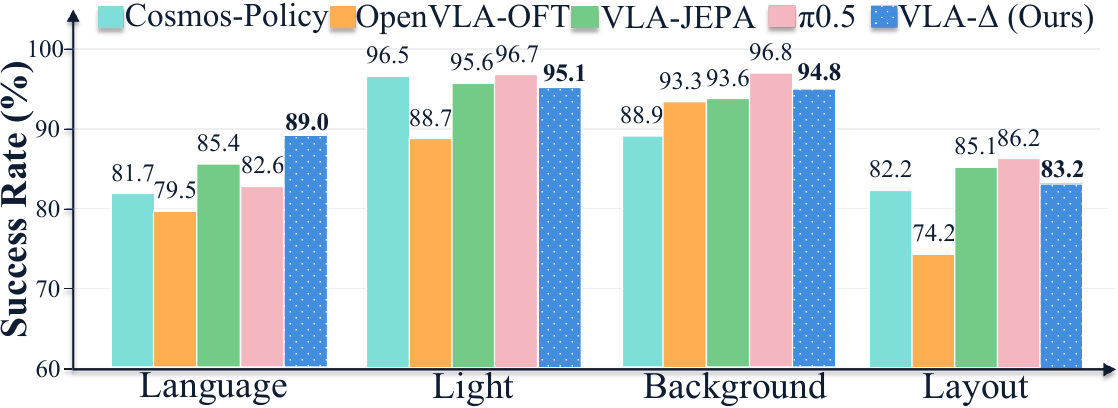}
    \captionsetup{hypcap=false}
    \captionof{figure}{LIBERO-Plus performance under four different perturbations.}
    \label{fig:libero-plus}
\end{minipage}
\par

As shown in Figure~\ref{fig:libero-plus}, \VLADelta{} maintains a high task success rate
across all perturbation settings and remains competitive with leading methods. These results indicate that \VLADelta{} achieves strong performance on the standard LIBERO
benchmark without substantially compromising its ability to adapt to changes in visual appearance,
background, and layout.

Notably, under the Language perturbation, \VLADelta{} achieves
\textbf{89.0\%}, outperforming $\pi_{0.5}$ at \textbf{82.6\%} by \textbf{6.4 percentage points}.
This result suggests that the explicit semantic information provided by the Future-State Difference
Text enables the policy to rely more on task-relevant state transitions when interpreting
instructions, thereby reducing its sensitivity to superficial variations in language.

\input{tables/table2_robot_task_results}

\paragraph{\textbf{RoboTwin 2.0.}} We further evaluate \VLADelta{} on RoboTwin 2.0. As shown in Table~\ref{tab:robot_task_results},
\VLADelta{} achieves average success rates of 90.06\% and 89.74\% under the Clean and Random
settings, respectively, outperforming all compared methods in both settings. These results exceed
those of ZR-0, the strongest baseline in both settings, by 1.36 and 1.76 percentage points. From
Clean to Random, the average success rate of \VLADelta{} decreases by only 0.32 percentage
points, compared with 0.72 for ZR-0 and 1.50 for Motus. This indicates that the model achieves
higher success rates while adapting effectively to scene randomization.

Under the Clean setting, \VLADelta{} achieves the highest success rates on BlocksRankingSize and
HangingMug among all compared methods. BlocksRankingSize requires arranging three blocks by size
through repeated grasping, transport, and placement, with successive operations adjusting spatial
relations according to earlier placement outcomes. HangingMug involves rotation, regrasping, and
hanging, requiring coordinated changes in object poses and transitions among tabletop support,
gripper grasping, and rack support. The state reached at each stage influences the conditions for
subsequent operations. These two tasks therefore emphasize progressive spatial rearrangement and
contact transitions across multiple stages, respectively. The observed gains support our central
premise that explicitly learning changes in task progress, spatial motion, and interaction states
helps connect task semantics with continuous action generation.

Overall, \VLADelta{} outperforms the compared approaches that use intermediate reasoning or
future visual state prediction on downstream tasks. These results suggest that treating
task-relevant motion and interaction changes as explicit learning targets guides action generation
more effectively.

\subsection{Real-Robot Evaluation}

\begin{figure}[t]
    \centering
    \includegraphics[width=\textwidth,height=0.82\textheight,keepaspectratio]{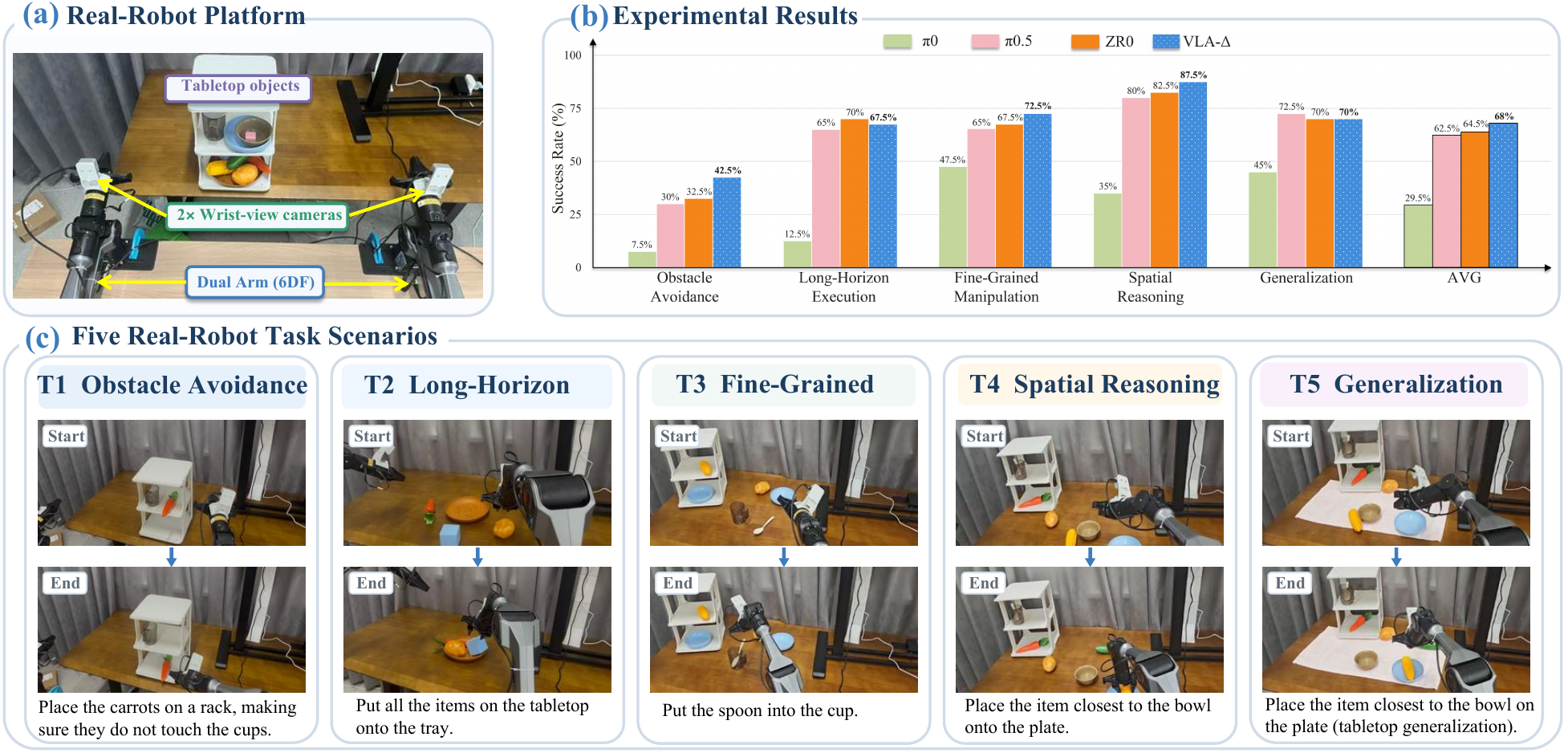}
    \caption{Real-robot evaluation on the AgileX PiPER 6-DoF platform. (a) Dual-arm robot
    platform. (b) Real-robot experimental results (Success Rates, \%). (c) Overview of the five
    tasks.}
    \label{fig:real-robot-evaluation}
\end{figure}

Figure~\ref{fig:real-robot-evaluation} presents the real-robot platform, task scenarios, and
experimental results. We design five manipulation tasks to evaluate obstacle avoidance,
long-horizon execution, fine-grained manipulation, spatial reasoning, and generalization under
environmental changes. Each task is evaluated over 40 independent episodes, for a total of
200 trials.

Task success rates are reported in Figure~\ref{fig:real-robot-evaluation}(b). \VLADelta{} achieves
an average success rate of 68.0\% across the five tasks. On the obstacle avoidance task, it
achieves 42.5\%, the highest success rate among all compared methods. This result is consistent
with the explicit Object Constraints supervision in Future-State Difference Text. These
annotations describe obstacles to avoid and object relationships to preserve, helping the model
account for manipulation constraints and potential collision risks during action generation.

\VLADelta{} also achieves the highest success rates among compared methods on the spatial
reasoning and fine-grained manipulation tasks. This advantage may reflect spatial position
annotations in Structured Transition Slots and visual details from optical flow. On
the long-horizon execution and environmental generalization tasks, \VLADelta{} achieves the
second-highest success rates. These results indicate continued effectiveness on complex
long-horizon tasks and under environmental changes.

Overall, the experimental results support the effectiveness of multilevel physical
state-transition supervision in helping policies learn task-relevant state changes and their
constraints, and translate this information into continuous control in real-world environments.

\subsection{Ablation Experiments}

To analyze the contributions of Physical State Difference Queries, Future-State Difference Text,
Structured Transition Slots, and Dense Optical Flow, we conduct ablation experiments on LIBERO.
All variants use only the action-learning objective during downstream fine-tuning.

\input{tables/table4_ablation_libero}

As shown in Table~\ref{tab:ablation_libero}, compared with the baseline using dense vision-language
features, introducing the query mechanism
improves the success rate by 0.55 percentage points, indicating that compact query representations
support action generation. To distinguish the effects of pretraining data from those of auxiliary
supervision, we further pretrain both configurations using only action supervision, which yields
limited performance gains.

\begin{figure}[t]
    \centering
    \includegraphics[width=\linewidth]{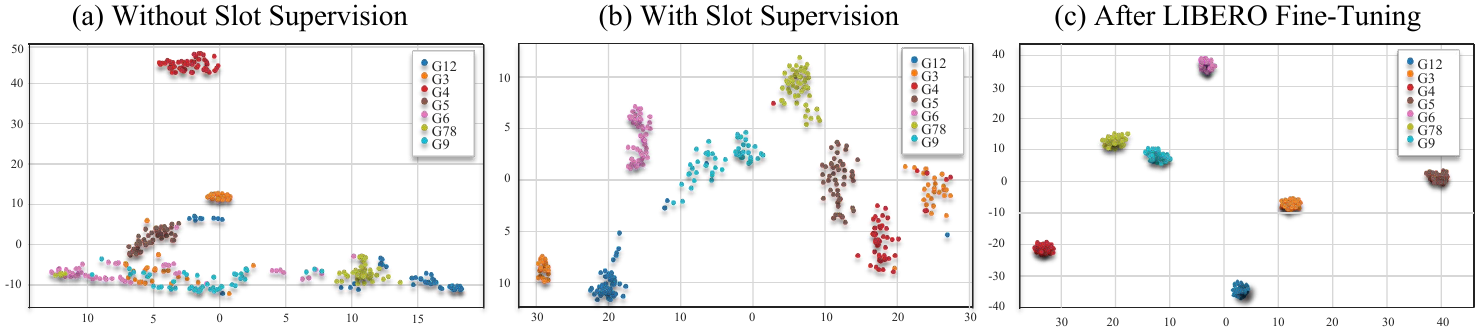}
    \captionsetup{justification=centering}
    \caption{t-SNE visualization of Slot Query hidden states.}
    \label{fig:ablation-visualization}
\end{figure}

We then introduce Future-State Difference Text, Structured Transition Slots, and Dense Optical
Flow separately, and all three configurations yield further performance improvements. We
investigate why Structured Transition Slots improve model performance by extracting the hidden
states of Slot Queries from models pretrained with and without structured slot supervision and
projecting them into two dimensions using t-SNE. The nine slot targets use seven fixed Query groups.
Task progress and phase share G12, while gripper state and contact transitions share G78, with
separate output branches for each target. Appendix~\ref{sec:slot-head-details} details the grouping,
prediction heads, and supervision objectives.
As shown in Figures~\ref{fig:ablation-visualization}(a)
and~\ref{fig:ablation-visualization}(b), query representations are relatively dispersed without
structured supervision during pretraining. With Structured Transition Slots, the query groups form
more compact clusters with clearer boundaries. This observation suggests that structured slot
supervision enhances the semantic separability of query representations.

\noindent
\begin{minipage}[t]{0.48\textwidth}
    \vspace{0pt}
    To examine whether this structure transfers to downstream tasks, we extract the Slot Query hidden
    states after task-specific fine-tuning on LIBERO. As shown in
    Figure~\ref{fig:ablation-visualization}(c), the Slot Queries still form seven compact,
    well-separated clusters, although fine-tuning uses only the action-generation objective.
    This indicates that the query grouping structure established during pretraining
    persists through downstream fine-tuning, suggesting that the semantic prior introduced by
    Structured Transition Slots can continue to guide the learning of task-relevant transition
    representations.

    Finally, the complete \VLADelta{} model combines all three supervision types, achieving an
    average success rate of \textbf{98.4\%}. This result suggests that multilevel physical
    state-transition supervision jointly shapes more comprehensive transition representations,
    improving task execution.

\end{minipage}\hfill
\begin{minipage}[t]{0.48\textwidth}
    \vspace{0pt}
    \centering
    \includegraphics[width=\linewidth]{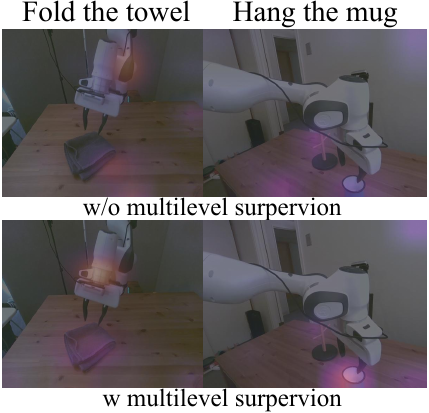}
    \captionsetup{hypcap=false}
    \captionof{figure}{Query attention maps for towel folding and mug hanging.}
    \label{fig:attention-heatmap}
\end{minipage}
\par

We also visualize query attention distributions over the input images for the complete
\VLADelta{} model and the variant without the three transition-supervision signals. As shown in
Figure~\ref{fig:attention-heatmap}, queries trained with all three types of supervision attend
more strongly to the target object, gripper, contact regions, and other task-relevant objects.
In contrast, the model without transition supervision exhibits more diffuse attention and
assigns greater attention to task-irrelevant background regions.

%% file: tables/table1_libero.tex
\begin{table}[!t]
    \centering
    \captionsetup{justification=centering}
    \caption{Performance comparison on LIBERO (Success Rates,\%).}
    \label{tab:libero_main}
    \begingroup
    \scriptsize
    \begin{tabularx}{\textwidth}{l*{6}{>{\centering\arraybackslash}X}}
        \toprule
        \textbf{Methods} & \textbf{VLM Size} & \textbf{Spatial} & \textbf{Object}
        & \textbf{Goal} & \textbf{Long} & \textbf{Average} \\
        \midrule
        TraceVLA      & 7B & 84.6 & 85.2 & 75.1 & 54.1 & 74.8 \\
        openVLA       & 7B & 84.7 & 88.4 & 79.2 & 53.7 & 76.5 \\
        SpatialVLA    & 4B & 88.2 & 89.9 & 78.6 & 55.5 & 78.1 \\
        WorldVLA      & 7B & 87.6 & 96.2 & 83.4 & 60.0 & 81.8 \\
        CoT-VLA       & 7B & 87.5 & 91.6 & 87.6 & 69.0 & 83.9 \\
        ThinkAct      & 7B & 88.3 & 91.4 & 87.1 & 70.9 & 84.4 \\
        SmolVLA       & 500M & 93.0 & 94.0 & 91.0 & 77.0 & 88.8 \\
        DreamVLA      & 345M & 97.5 & 94.0 & 89.5 & 89.5 & 92.6 \\
        $\pi_0$       & 3B & 96.8 & 98.8 & 95.8 & 85.2 & 94.2 \\
        UniVLA        & 7B & 95.4 & 98.8 & 93.6 & 94.0 & 95.5 \\
        $\pi_{0.5}$   & 3B & $\mathbf{98.8}$ & 98.2 & 98.0 & 92.4 & 96.9 \\
        GR00T-N1.7    & 2B & 97.7 & 97.5 & \underline{98.5} & 94.4 & 97.0 \\
        openVLA-OFT   & 7B & 97.6 & 98.4 & 97.9 & 94.6 & 97.1 \\
        VLA-Adapter   & 0.5B & \underline{97.8} & 99.2 & 97.2 & \underline{95.0} & 97.3 \\
        ZR-0          & 2B & 97.4 & \underline{99.4} & 97.8 & $\mathbf{96.4}$ & \underline{97.8} \\
        \midrule
        \rowcolor{blue!10}
        \textbf{VLA-$\Delta$}
                      & \textbf{2B}
                      & $\mathbf{98.8}$
                      & $\mathbf{99.8}$
                      & $\mathbf{98.6}$
                      & $\mathbf{96.4}$
                      & $\mathbf{98.4}$ \\
        \bottomrule
    \end{tabularx}
    \endgroup
\end{table}

%% file: tables/table2_robot_task_results.tex
\begin{table*}[t]
    \centering
    \caption{Performance comparison on RoboTwin 2.0 (Success Rates, \%) under Clean and Random
    settings. Full results are provided in Appendix~\ref{sec:appendix-robotwin}.}
    \label{tab:robot_task_results}
    \begingroup
    \setlength{\tabcolsep}{3.5pt}
    \resizebox{\textwidth}{!}{%
    \begin{tabular}{lccccccccccccc}
        \toprule
        \multirow{2}{*}{\textbf{Method}}
        & \multicolumn{2}{c}{\textbf{AdjustBottle}}
        & \multicolumn{2}{c}{\textbf{BeatBlockHammer}}
        & \multicolumn{2}{c}{\textbf{BlocksRankingSize}}
        & \multicolumn{2}{c}{\textbf{ClickAlarmclock}}
        & \multicolumn{2}{c}{\textbf{HangingMug}}
        & \multirow{2}{*}{\textbf{\ldots}}
        & \multicolumn{2}{c}{\textbf{AVG}} \\
        \cmidrule(lr){2-3}
        \cmidrule(lr){4-5}
        \cmidrule(lr){6-7}
        \cmidrule(lr){8-9}
        \cmidrule(lr){10-11}
        \cmidrule(lr){13-14}
        & \textbf{Clean} & \textbf{Random}
        & \textbf{Clean} & \textbf{Random}
        & \textbf{Clean} & \textbf{Random}
        & \textbf{Clean} & \textbf{Random}
        & \textbf{Clean} & \textbf{Random}
        &
        & \textbf{Clean} & \textbf{Random} \\
        \midrule
        X-VLA
        & \textbf{100} & \textbf{99} & 92 & \underline{88} & 67 & \underline{74}
        & \underline{99} & \underline{99} & 23 & 27
        & \ldots
        & 72.88 & 72.84 \\
        $\pi_{0}$
        & \underline{99} & 95 & 79 & 84 & 14 & 5 & 77 & 68 & 14 & 11
        & \ldots
        & 65.92 & 58.4 \\
        $\pi_{0.5}$
        & \textbf{100} & \textbf{99}
        & \textbf{96} & \textbf{93}
        & 49 & 26
        & 98 & 89
        & 18 & 17
        & \ldots
        & 82.74 & 76.76 \\
        Motus
        & 89 & 93 & \underline{95} & \underline{88} & \underline{75} & 63
        & \textbf{100} & \textbf{100} & \underline{38} & \textbf{38}
        & \ldots
        & 88.52 & 87.02 \\
        ZR-0
        & \textbf{100} & \textbf{99} & 85 & \underline{88} & 70 & \textbf{81}
        & 96 & 82 & 35 & 33
        & \ldots
        & \underline{88.70} & \underline{87.98} \\
        \midrule
        \rowcolor{blue!10}
        \textbf{VLA-$\Delta$}
        & \textbf{100} & \underline{98} & 90 & 86 & \textbf{80} & \underline{74}
        & 92 & 94 & \textbf{40} & \underline{34}
        & \ldots
        & \textbf{90.06} & \textbf{89.74} \\
        \bottomrule
    \end{tabular}%
    }
    \endgroup
\end{table*}

%% file: tables/table4_ablation_libero.tex
\definecolor{darkgreen}{RGB}{0,100,0}

\begin{table}[t]
    \centering
    \captionsetup{justification=centering}
    \caption{Ablation results on 40 LIBERO tasks.}
    \label{tab:ablation_libero}
    \begingroup
    \scriptsize
    \setlength{\tabcolsep}{4.8pt}
    \renewcommand{\tabularxcolumn}[1]{m{#1}}
    \begin{tabularx}{\textwidth}{l*{6}{>{\centering\arraybackslash}X}}
        \toprule
        \multicolumn{1}{c}{\textbf{Variant}}
        & \textbf{Query}
        & \textbf{Pretraining}
        & \textbf{Difference Text}
        & \textbf{Slots}
        & \textbf{Optical Flow}
        & \textbf{LIBERO Avg. (\%)} \\
        \midrule
        Dense VLM Features
        & \textcolor{red}{No} & \textcolor{red}{No} & \textcolor{red}{No}
        & \textcolor{red}{No} & \textcolor{red}{No}
        & 95.70 \\

        Dense VLM Features
        & \textcolor{red}{No} & \textcolor{darkgreen}{Yes} & \textcolor{red}{No}
        & \textcolor{red}{No} & \textcolor{red}{No}
        & 95.85 \\

        Difference Queries Only
        & \textcolor{darkgreen}{Yes} & \textcolor{red}{No} & \textcolor{red}{No}
        & \textcolor{red}{No} & \textcolor{red}{No}
        & 96.25 \\

        Difference Queries Only
        & \textcolor{darkgreen}{Yes} & \textcolor{darkgreen}{Yes} & \textcolor{red}{No}
        & \textcolor{red}{No} & \textcolor{red}{No}
        & 96.45 \\

        Query + Difference Text
        & \textcolor{darkgreen}{Yes} & \textcolor{darkgreen}{Yes} & \textcolor{darkgreen}{Yes}
        & \textcolor{red}{No} & \textcolor{red}{No}
        & 97.20 \\

        Query + Structured Slots
        & \textcolor{darkgreen}{Yes} & \textcolor{darkgreen}{Yes} & \textcolor{red}{No}
        & \textcolor{darkgreen}{Yes} & \textcolor{red}{No}
        & 96.90 \\

        Query + Optical Flow
        & \textcolor{darkgreen}{Yes} & \textcolor{darkgreen}{Yes} & \textcolor{red}{No}
        & \textcolor{red}{No} & \textcolor{darkgreen}{Yes}
        & 97.35 \\

        \midrule

        \rowcolor{blue!10}
        \textbf{VLA-$\Delta$ (Full)}
        & \textcolor{darkgreen}{Yes} & \textcolor{darkgreen}{Yes} & \textcolor{darkgreen}{Yes}
        & \textcolor{darkgreen}{Yes} & \textcolor{darkgreen}{Yes}
        & \textbf{98.40} \\
        \bottomrule
    \end{tabularx}
    \endgroup
\end{table}

%% file: sections/conclusion.tex
\section{Conclusion}

We propose \VLADelta{}, a general framework that uses physical changes between current and
future states as explicit learning targets. Using robot demonstrations from multiple sources,
we develop an automated annotation pipeline to construct Future-State Difference Text,
Structured Transition Slots, and Optical Flow annotations, supervising transition representations
through task semantics, structured state variables, and dense visual motion. These supervision
signals are used only during pretraining to learn representations that support downstream policy
learning through action-only fine-tuning, while deployment requires no decoding of intermediate
supervision targets. Experiments demonstrate that \VLADelta{} achieves competitive performance
on LIBERO and RoboTwin and the highest average success rate among the compared methods in
real-robot experiments. Ablation studies further validate the contributions of individual
supervision signals, indicating that explicitly learning physical state changes provides an
effective representational foundation for downstream continuous action generation.

%% file: sections/appendix.tex
\beginappendix

\input{sections/related_work}

\input{sections/appendix_pretraining}

\input{sections/appendix_evaluation}

\input{sections/appendix_training}

%% file: sections/related_work.tex
\section{Related Work}

\subsection{Vision-Language-Action Models}

Vision-language-action (VLA) models combine pretrained vision-language models with robot
demonstrations to connect multimodal understanding with action generation. Related research
provides complementary foundations for generalist robot policy learning
\citep{brohan2022rt,li2024cogact,wen2025dexvla,qu2025spatialvla,
li2024roboflamingo,ghosh2024octo,bousmalis2023robocat,li2024robovlms,
bu2025agibot,deng2025graspvla,bu2025univla,wu2026pragmatic,shukor2025smolvla,intelligence2026pi}.
RT-2 \citep{RT2023} encodes actions as discrete tokens and jointly trains on robot trajectories
and vision-language tasks. This formulation enables semantic knowledge from visual and linguistic
pretraining to support robot control. OpenVLA
\citep{kim2024openvla} trains an open generalist policy on large-scale robot demonstrations,
demonstrating the value of vision-language pretraining for multitask manipulation and policy adaptation.
For continuous control, $\pi_0$ and $\pi_{0.5}$ \citep{Pi02024,intelligence2025pi_} combine
vision-language pretraining with flow matching to generate continuous action chunks. GR00T N1
\citep{GR00T2025} combines a vision-language backbone with a flow-matching action expert,
providing the architectural basis for our model. Action representations and generative policy
learning provide complementary foundations for continuous control
\citep{peebles2023scalable,lipman2022flow,chi2023diffusionpolicy,zhao2023act,liu2024rdt,
kim2025oft,pertsch2025fast,yang2026abot}.

\VLADelta{} focuses on explicit supervision of physical changes between current and future
states to shape transition representations for continuous action generation. These supervision
signals are used only during pretraining, while downstream fine-tuning relies on action supervision.

\subsection{Embodied Reasoning}

Embodied reasoning links high-level task semantics with action execution through intermediate
plans, spatial reasoning, and grounded representations
\citep{ahn2022saycan,belkhale2024rth,huang2023voxposer,yuan2024robopoint,driess2023palm,
li2024manipllm,huang2024rekep,duan2024manipulate,fang2026molmoact2,zheng2025tracevla,shi2026memoryvla}.
ECoT \citep{ECoT2024} generates task plans, subgoals, and motion descriptions before action
prediction, grounding reasoning in the robot state. ThinkAct \citep{huang2025thinkact} uses
goal-completion and trajectory-consistency rewards to learn embodied reasoning and compresses
plans into latent representations that guide action policies. ZR-0 \citep{Li2026ZR0} learns
cross-embodiment representations through dense embodied chain-of-thought supervision and can
skip reasoning generation during deployment. ThinkingVLA \citep{lu2026thinkingvla} interleaves
forward textual reasoning, future-image prediction, and inverse reasoning within a unified
autoregressive architecture. Forward reasoning selects a subgoal and guides visual forecasting,
while inverse reasoning uses the predicted target state to infer spatial relations and action
intent. Actions are generated from the combined reasoning context.

These methods show how explicit reasoning can relate task goals and spatial information to robot control. \VLADelta{} focuses on
task-relevant physical changes between current and future states, combining Future-State
Difference Text, Structured Transition Slots, and optical flow to connect coherent semantic
descriptions with dense visual motion supervision.

\subsection{World Action Models}

In robotic manipulation, world action models connect predictions of future observations or latent
states with action generation. Broader work on predictive representations and policy learning
provides complementary ways to condition actions on anticipated outcomes
\citep{zhen2024threedvla,cheang2024gr2,bharadhwaj2024gen2act,GR12023,CoTVLA2025,
zhang2025up,chen2026lawam,JEPAWAM2026,wen2024atm,
bharadhwaj2024track2act,yuan2024generalflow}.
WorldVLA \citep{WorldVLA2025} unifies action generation and future-image generation within an
autoregressive framework, allowing information sharing between policy learning and world modeling.
VLA-JEPA \citep{VLAJEPA2026} predicts latent representations of future observations, learning
transition information while reducing sensitivity to appearance and background variations.
DreamVLA \citep{zhang2026dreamvla} forecasts dynamic, spatial, and semantic information to guide
action prediction. Motus \citep{bi2026motus} unifies understanding, video generation, and action
experts, using optical flow to learn latent actions. These approaches differ in whether future
information is represented by reconstructed images, latent features, or complementary prediction
targets, but all use predictive learning to support action generation beyond the current
observation. Generating future images online, however, adds visual-token decoding or video-denoising costs.

\VLADelta{} combines readable descriptions of task progress, spatial and contact changes, and
physical constraints with structured transition slots and dense optical flow supervision.
This joint supervision shapes a shared transition representation for action generation without
requiring future-image or difference-text generation at deployment.

%% file: sections/appendix_pretraining.tex
\section{Pretraining Dataset and Data Annotations}
\label{sec:appendix-pretraining}

\subsection{Pretraining Dataset}

To learn visual, language, and physical state-transition representations for robot manipulation,
we construct a pretraining dataset by selecting demonstrations from DROID, MolmoAct, and RH20T. We
first convert all data sources into a unified trajectory format, then filter samples based on the
completeness of visual observations, language instructions, robot states, and action records. The
resulting dataset covers diverse tasks, scene layouts, robot configurations, and interaction types.

Table~\ref{tab:pretraining_datasets} summarizes the scale and collection characteristics of the
three source datasets.

\input{tables/table_pretraining_datasets}

DROID contains approximately 76,000 demonstration trajectories across 564 scenes and 84 task
categories, totaling approximately 350 hours of interaction. We retain demonstrations with
complete timestamps, valid visual frames, and continuous action records, prioritizing trajectories
that include approaching, grasping, transporting, placing, and releasing objects. These
demonstrations increase the diversity of environments, objects, and tasks in the pretraining data.

MolmoAct contains over 10,000 high-quality manipulation trajectories covering household and
tabletop settings. We extract visual observations, language instructions, robot states, and
actions, retaining trajectories that clearly reflect task progress and subgoal changes. These data
provide additional information about language conditioning, spatial relations, and multistage
manipulation.

RH20T contains over 110,000 contact-rich manipulation sequences spanning multiple robot platforms,
sensor configurations, and manipulation skills, with modalities including vision, force sensing,
and audio. We use its visual observations, proprioceptive states, and continuous action records;
the remaining modalities are not directly provided to the policy network. Grasping, pushing,
insertion, and release sequences in this dataset provide additional examples of contact
interactions, object state changes, and fine-grained manipulation.

The three data sources complement one another in scene diversity, task semantics, and physical
interaction. After format standardization and filtering, the pretraining data cover diverse spatial
layouts, motion patterns, and manipulation constraints, providing a foundation for learning
multilevel physical state-transition representations across tasks and robot configurations.

\subsection{Physical State-Transition Annotations}
\label{sec:appendix-state-transition-annotations}

\begin{figure}[tbp]
    \centering
    \includegraphics[width=\linewidth,keepaspectratio,pagebox=mediabox]{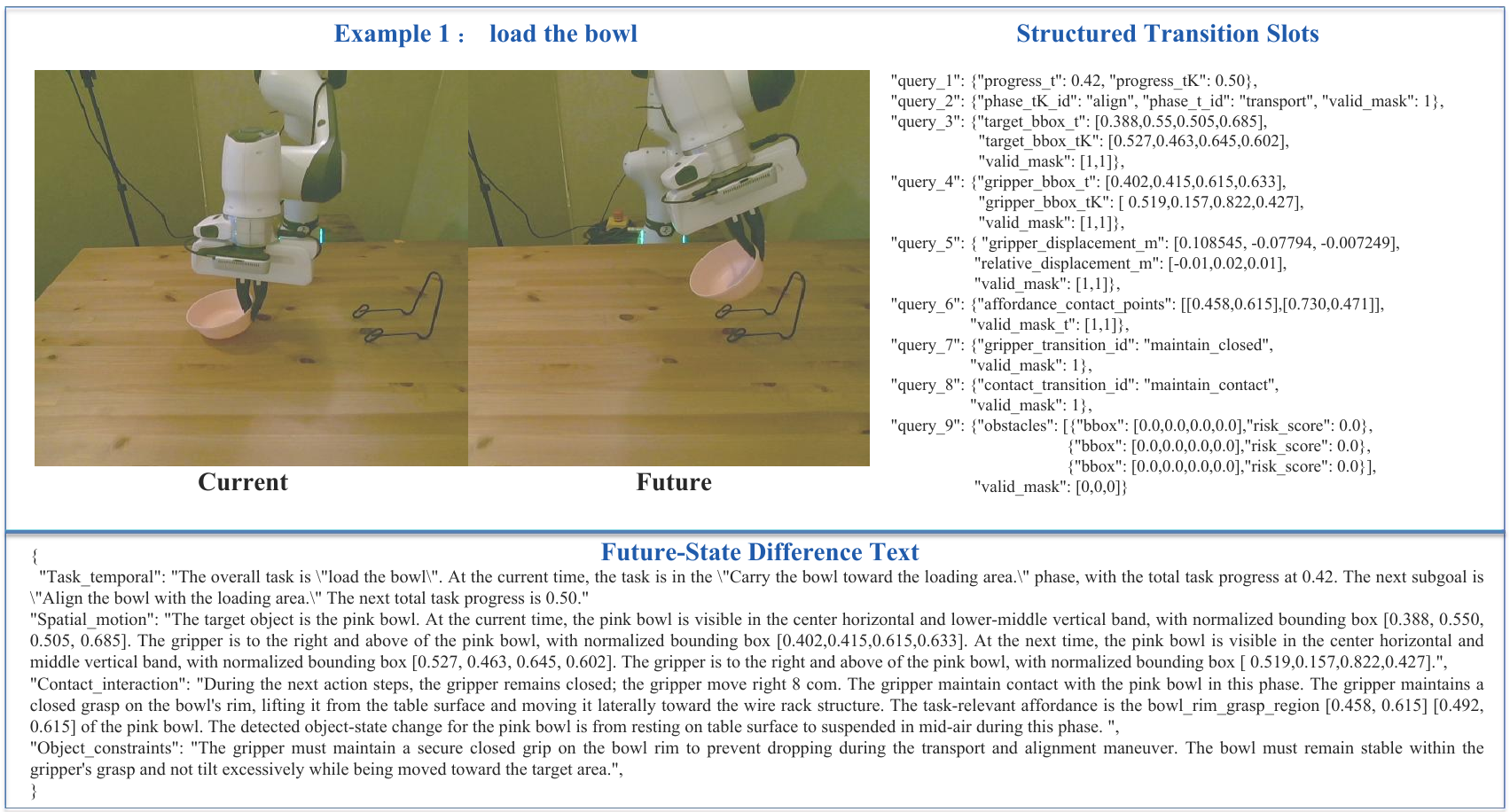}
    \par\smallskip
    \includegraphics[width=\linewidth,keepaspectratio]{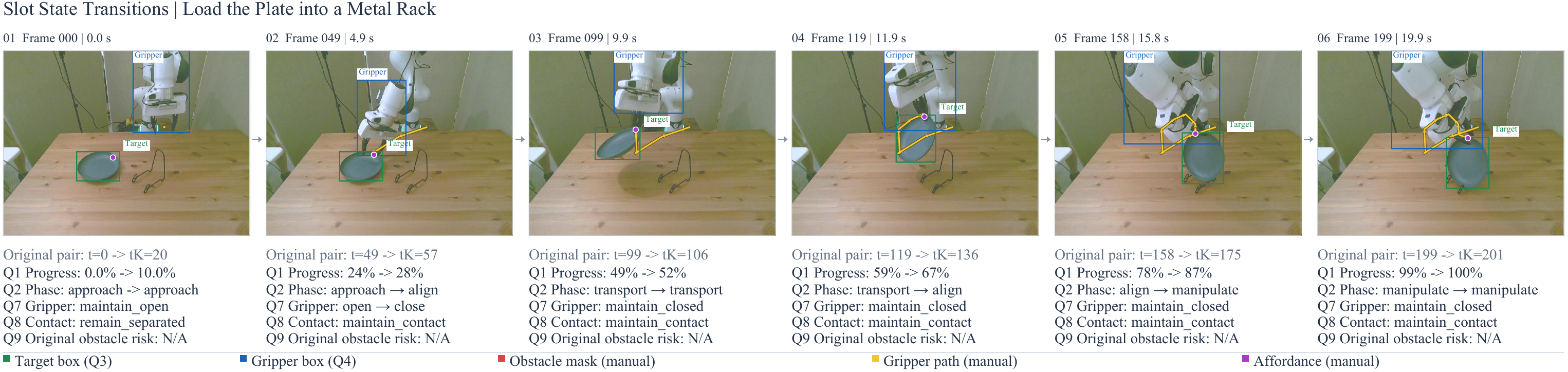}
    \par\smallskip
    \includegraphics[width=\linewidth,keepaspectratio]{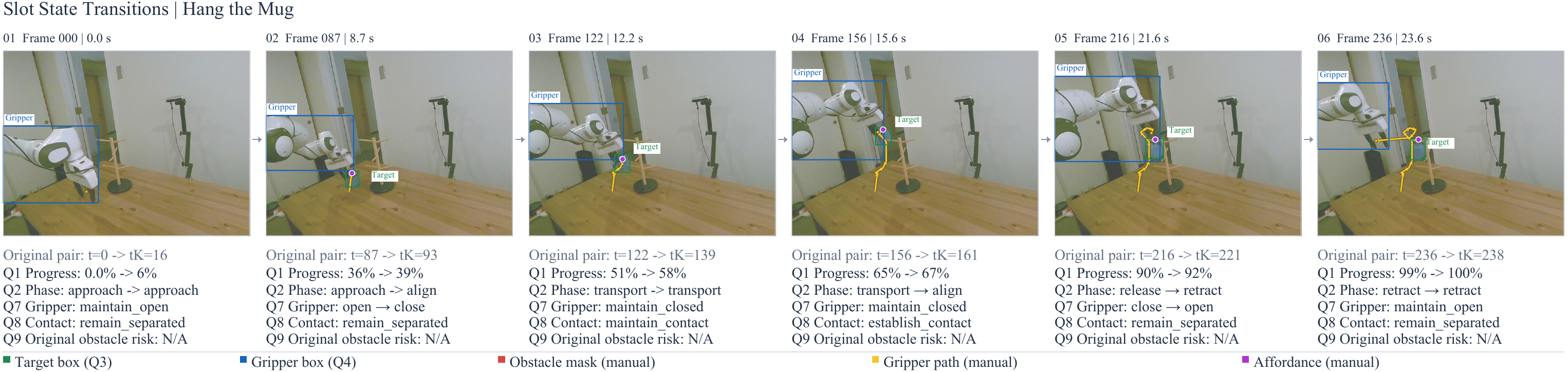}
    \caption{Physical state-transition annotation examples: a current--future observation pair with
    Future-State Difference Text and Structured Transition Slots (top), and slot annotations for
    loading a plate into a metal rack (middle) and hanging a mug (bottom).}
    \label{fig:annotation-examples}
\end{figure}

Figure~\ref{fig:annotation-examples} presents examples of physical state-transition annotations.

\FloatBarrier

\subsection{Automated Data Annotation Pipeline}
\label{sec:appendix-annotation-pipeline}

\paragraph{(1) Data Selection Strategy.}
We select demonstrations with clear goals, valid observations, complete state and action records,
and reliable temporal alignment, considering diversity in tasks, objects, scenes, robot
configurations, and viewpoints. Operations cover approaching, aligning, grasping, transporting,
placing, releasing, pushing, and insertion. For partially corrupted trajectories, we retain
continuous, semantically coherent segments and discard records with ambiguous target identities,
action meanings, or temporal correspondences.

\paragraph{(2) Unified Trajectory Format.}
We convert all sources to LeRobot v3, organizing instructions, multiview images, proprioceptive
states, actions, and timestamps. Time-series data are stored in Parquet and visual observations in
videos, with metadata specifying features, frame rates, and trajectory boundaries. We retain
source and trajectory identifiers, robot and camera configurations, calibration information, and
field dimensions, units, coordinate frames, and validity indicators. Modalities are aligned by
timestamps; missing or unreliably aligned fields are marked invalid.

\paragraph{(3) Data Cleaning.}
Language instruction checks. We identify empty, garbled, truncated, or inconsistent descriptions,
correcting recoverable instructions and excluding unrecoverable records from language-conditioned
training. Language standardization preserves original text and verifies semantic consistency.
Future-state difference descriptions remain supervision labels, separate from input instructions.

Temporal consistency checks. We inspect timestamp ordering, duplicates, missing frames, and
cross-modal offsets. Different video and control frequencies are permitted when reliably aligned.
Temporal gaps define segment boundaries.

Visual quality checks. We identify undecodable, black, overexposed, blurred, or uninformative views.
Brief occlusions are retained when neighboring frames or other views provide evidence; otherwise,
affected annotations or segments are excluded.

State and action checks. We inspect missing dimensions, non-finite values, jumps, and out-of-range
records using sampling intervals and robot configurations. Unclear action types, rotation
representations, units, or coordinate frames are not speculatively converted.

Trajectory integrity checks. We identify abnormal lengths, duplicates, and truncation,
distinguishing valid waiting and sustained contact from recording stalls. When applicable,
training and validation sets are split by trajectory before extracting windows to prevent leakage.

\paragraph{(4) Adaptive Keyframe Selection and Semantic Annotation.}
Qwen3.8-27B identifies operation phases from instructions and videos, selecting boundaries and
surrounding frames. Sampling is denser for precise alignment, insertion, gripper changes, and
contact transitions, and sparser for smooth transport, while preserving task-critical events.

Using keyframes and neighboring clips, the model annotates objects, phases, contact relations,
interaction regions, state changes, constraints, and obstacles. Annotations distinguish current
states, observed future changes, and goals, as well as explicit and inferred constraints. Fields
retain object identifiers, timestamps, supporting frames, and uncertainty indicators.

\paragraph{(5) Visual Localization, Depth, and Optical Flow Annotation.}
SAM 3 produces bounding boxes and masks for targets, grippers, and obstacles. Neighboring frames
and tracks maintain object identities through similar-instance ambiguity and occlusion.

UniDepth V2 estimates depth, summarized by the local median within valid target-mask regions near
the two-dimensional center. This measures visible-surface depth, not the object's geometric
center. Severe occlusion or unstable estimates invalidate the field.

MegaFlow estimates dense two-dimensional displacement offline between current and subsequent
frames from the same viewpoint. Frame identifiers and intervals maintain correspondence with
transition descriptions.

\paragraph{(6) Annotation Integration and Training Sample Construction.}
Shared records for current--future state pairs contain intervals, object identities, phases,
positions, contacts, changes, and constraints. They generate Future-State Difference Text and
Structured Transition Slots, providing semantic descriptions and structured categories, values,
and validity indicators. Optical flow supplies dense motion information, with all annotations
sharing temporal and object correspondences.

Keyframes locate events, while the model's prediction horizon determines future endpoints.
Information derived from future observations serves only as pretraining supervision, never as
input for current action prediction.

\paragraph{(7) Stratified Quality Review and Traceability.}
Programmatic checks verify completeness, timing, coordinates, identities, and cross-annotation
consistency. An independent multimodal model reviews difficult or conflicting samples field by
field against visual evidence, retaining uncertainty where necessary. Correctable errors are
reannotated, invalid fields are masked, and entire samples are discarded only when essential
timing or target identities cannot be established.

Manual inspections are stratified by source, task, phase, and quality level, with random checks
of automatically accepted samples to estimate missed errors. Findings guide revisions to prompts,
rules, and guidelines and reprocessing of affected data. Sources, model versions, evidence frames,
review decisions, and revision histories support traceability and reproducibility.

\paragraph{(8) Annotation Audit Statistics.}
We sample 300 current--future state pairs, with 100 each from DROID, MolmoAct, and RH20T,
and audit the four aspects of Future-State Difference Text and the nine Structured Transition
Slot categories. Table~\ref{tab:annotation_quality_audit} summarizes the reference criteria,
valid counts, and corresponding proportions of the sampled pairs for each category.

\input{tables/table_annotation_quality_audit}

%% file: tables/table_pretraining_datasets.tex
\begin{table}[htbp]
    \centering
    \caption{Overview of the datasets used to construct the pretraining corpus.}
    \label{tab:pretraining_datasets}
    \small
    \setlength{\tabcolsep}{3pt}
    \renewcommand{\arraystretch}{1.12}
    \begin{tabularx}{\textwidth}{@{}>{\raggedright\arraybackslash}p{0.18\textwidth}rrr>{\raggedright\arraybackslash}X@{}}
        \toprule
        \textbf{Dataset} & \textbf{Frames} & \textbf{Hours} & \shortstack{\textbf{No. of}\\\textbf{Tasks}}
        & \textbf{Introduction} \\
        \midrule
        MolmoAct (Household + Tabletop)
        & 1,104,942 & 30.693 & 96
        & Everyday household and tabletop manipulation with a Franka robot, covering organizing,
        cleaning, opening and closing, grasping, and placement tasks; includes a fixed external
        view, a second external view, and a wrist-camera view. \\
        \addlinespace
        DROID
        & 27,630,375 & 511.674 & 49,630
        & Open-ended everyday manipulation with Franka robots across different data collectors,
        buildings, and real-world scenes; includes two external views and a wrist-camera view. \\
        \addlinespace
        RH20T
        & 3,898,056 & 108.279 & 141
        & Multi-robot real-world manipulation. The Stage05 release subset covers Flexiv, Franka,
        KUKA, and UR5 across seven configurations, 19 users, and 21 scene IDs; includes two
        external views, a wrist-camera view, and force/torque fields. \\
        \bottomrule
    \end{tabularx}
\end{table}

%% file: tables/table_annotation_quality_audit.tex
\begin{table}[t]
    \centering
    \caption{Counts and proportions from the audit of physical state-transition annotations.}
    \label{tab:annotation_quality_audit}
    \small
    \begin{tabularx}{\linewidth}{@{}>{\raggedright\arraybackslash}p{0.28\linewidth}>{\raggedright\arraybackslash}Xrr@{}}
        \toprule
        Evaluation target & Reference criteria & Valid count & Percentage \\
        \midrule
        Text: Task Temporal
        & Human-verified task phases, task progress, and subgoal changes
        & 290 & 96.67\% \\
        Text: Spatial Motion
        & Human-verified object and gripper positions, relative relations, and their changes
        & 289 & 96.33\% \\
        Text: Contact Interaction
        & Gripper opening and closing, contact establishment, maintenance, and release confirmed in video
        & 293 & 97.67\% \\
        Text: Object Constraints
        & Manipulation constraints specified by task instructions or supported by visual evidence
        & 298 & 99.33\% \\
        \midrule
        Slot 1: Task progress
        & Human reference values in $[0,1]$ based on a consistent definition of progress
        & 292 & 97.33\% \\
        Slot 2: Task phase
        & Human labels based on predefined phase categories
        & 293 & 97.67\% \\
        Slot 3: Target position
        & Target-object bounding boxes annotated by GPT-5.6
        & 298 & 99.33\% \\
        Slot 4: Gripper position
        & Gripper bounding boxes annotated by GPT-5.6
        & 288 & 96.00\% \\
        Slot 5: Motion and relative displacement
        & Reference displacements from manually matched points or reliable calibration, in the same coordinate frame as the predictions
        & 279 & 93.00\% \\
        Slot 6: Interactable position
        & Human-annotated regions for grasping, contact, or insertion
        & 294 & 98.00\% \\
        Slot 7: Gripper state transitions
        & Human-verified gripper opening and closing transition labels
        & 298 & 99.33\% \\
        Slot 8: Contact transitions
        & Human-verified labels for contact establishment, maintenance, and release
        & 296 & 98.67\% \\
        Slot 9: Obstacles and risk
        & Human-annotated obstacle boxes and risk labels based on consistent risk criteria
        & 287 & 95.67\% \\
        \bottomrule
    \end{tabularx}
    \par\smallskip
    \begin{minipage}{\linewidth}
        \footnotesize
        Note: The audit includes 300 current--future state pairs, with 100 each from DROID,
        MolmoAct, and RH20T. Percentages equal the valid count divided by 300, rounded to
        two decimal places. Rows assess different aspects of the same samples, so their counts
        are not additive. Reference boxes for Slots 3 and 4 are generated by GPT-5.6 and do not
        constitute independent human ground truth. These percentages do not represent bounding-box
        intersection over union or displacement error.
    \end{minipage}
\end{table}

%% file: sections/appendix_evaluation.tex
\section{Evaluation Protocol}
\label{sec:appendix-evaluation-protocol}

\subsection{LIBERO and LIBERO-Plus}

\textbf{Benchmark overview.}
LIBERO is a simulation benchmark for evaluating knowledge transfer and multitask manipulation in
robotics. It includes four task suites, Spatial, Object, Goal, and Long, which assess spatial
understanding, object manipulation, goal-conditioned task completion, and long-horizon execution,
respectively. The benchmark provides diverse tabletop manipulation tasks with language
instructions, covering grasping, placing, moving, stacking, opening, and combinations of these
operations.

\textbf{The four suites emphasize different capabilities.}
\begin{itemize}
    \item Spatial evaluates the model's understanding of relative positions, orientations, and
    spatial relations, such as placing an object to the left or right of another.
    \item Object evaluates target identification and object recognition, requiring the policy to
    select the correct object and perform the specified operation.
    \item Goal assesses whether the model achieves the target state specified by a language
    instruction, such as placing an object inside a container or moving it to a designated location.
    \item Long comprises sequences of subgoals, requiring the model to track task progress,
    execution history, and intermediate states over extended horizons.
\end{itemize}

\textbf{Training data and fine-tuning.}
We perform multitask training using demonstrations from all four LIBERO task suites, with a total
of 1,693 training trajectories.

\textbf{Standard LIBERO evaluation.}
We evaluate the model on each of the four suites and report both suite-level and overall average
success rates. Each task is evaluated over 50 independent episodes, yielding 2,000 evaluation
episodes in total. Each episode uses a predefined initial scene configuration and random state.
The model performs closed-loop control using only the current observations, task instruction,
and robot state.

\textbf{LIBERO-Plus generalization evaluation.}
To assess adaptation to out-of-distribution environmental changes, we further evaluate the model
on LIBERO-Plus. We select four perturbation types and evaluate them across all four task suites,
running 50 episodes per task under each perturbation type for a total of 8,000 episodes. These
perturbations preserve the task goals and underlying manipulation logic while modifying the
instructions or environment.

\begin{itemize}
    \item Language varies instruction wording, sentence structure, or phrasing.
    \item Light changes lighting direction, brightness, or color.
    \item Background modifies background objects, tabletop textures, or scene appearance.
    \item Layout changes initial object configurations, relative positions, or scene arrangements.
\end{itemize}

We use no additional demonstrations from perturbed environments and perform no
perturbation-specific fine-tuning for LIBERO-Plus. The model retains the parameters learned on
standard LIBERO and uses the same input interface, action space, and maximum execution steps.

\subsection{RoboTwin 2.0}
\label{sec:appendix-robotwin}

\textbf{Benchmark overview.}
RoboTwin 2.0 is a simulation benchmark for dual-arm robot manipulation, focusing on bimanual
coordination, object interaction, spatial manipulation, and policy generalization under
environmental changes. It includes 50 dual-arm manipulation tasks and supports multiple robot
configurations and scene randomization. Scene generation covers diverse object categories and
instances, tabletop layouts, backgrounds, lighting conditions, and language instructions. The
official dataset provides a large object library and multiple structured domain randomization
settings.

\textbf{RoboTwin 2.0 provides two main settings.}
\begin{itemize}
    \item Clean uses relatively fixed scene appearances and object layouts to evaluate basic
    dual-arm manipulation in standard environments.
    \item Random introduces variations in object instances, initial positions, lighting,
    backgrounds, tabletop layouts, and instruction phrasing to assess generalization under
    distribution shifts.
\end{itemize}

The tasks involve bimanual grasping, coordinated transport, object arrangement, container
manipulation, tool use, and multi-object organization, placing substantial demands on bimanual
coordination, temporal planning, and contact control.

\textbf{Training data.}
We train on 27,500 demonstration trajectories across all 50 tasks. Each task contributes 50 Clean
demonstrations and 500 Random demonstrations.

\textbf{Evaluation protocol.}
We evaluate all 50 tasks under both Clean and Random settings, running 100 episodes per task in
each setting for a total of 10,000 evaluation episodes. During inference, the action chunk length
is set to 16. The policy executes the first eight actions of each predicted chunk before
generating a new chunk from updated observations, enabling closed-loop control.

Table~\ref{tab:robotwin_full_results} reports the complete per-task results under the Clean and
Random settings of RoboTwin 2.0.

\clearpage
\input{tables/table5_robotwin_full}
\clearpage

\subsection{Real-Robot Evaluation}
\label{sec:appendix-real-robot}

\textbf{Experimental platform.}
To evaluate the effectiveness of \VLADelta{} in real-world environments, we conduct experiments
on the AgileX PiPER 6-DoF dual-arm robot platform. The platform comprises left and right robotic
arms with end effectors, three cameras (one head camera and two wrist cameras), and robot
proprioceptive state feedback.

\textbf{Task setup.}
We design five tasks to evaluate physical constraint compliance, long-horizon execution,
fine-grained manipulation, spatial-relation reasoning, and generalization under environmental
changes, respectively.

The first four tasks are performed in a standard tabletop environment. The fifth task keeps the
language instruction and task goal unchanged while modifying only the tablecloth or tabletop
appearance to evaluate generalization to visual background changes.

\input{tables/table_real_robot_tasks}
\FloatBarrier

\textbf{Evaluation protocol.}
Each task is evaluated over 40 independent episodes, for a total of 200 real-robot trials.
For each task, all episodes use the same task instruction and a predefined distribution of
initial scene configurations, while specific object positions, poses, or background conditions
may vary according to the experimental setup.

%% file: tables/table5_robotwin_full.tex
\begingroup
\scriptsize
\setlength{\tabcolsep}{2.15pt}
\setlength{\LTcapwidth}{\textwidth}
\renewcommand{\arraystretch}{0.94}
\begin{longtable}{@{}l*{12}{r}@{}}
    \caption{Per-task success rates (\%) on RoboTwin 2.0 under the Clean and Random settings.}
    \label{tab:robotwin_full_results} \\
    \toprule
    \multirow{2}{*}{\textbf{Task}}
    & \multicolumn{2}{c}{$\boldsymbol{\pi_0}$}
    & \multicolumn{2}{c}{$\boldsymbol{\pi_{0.5}}$}
    & \multicolumn{2}{c}{\textbf{X-VLA}}
    & \multicolumn{2}{c}{\textbf{Motus}}
    & \multicolumn{2}{c}{\textbf{ZR-0}}
    & \multicolumn{2}{c}{\textbf{\VLADelta{}}} \\
    \cmidrule(lr){2-3}\cmidrule(lr){4-5}\cmidrule(lr){6-7}
    \cmidrule(lr){8-9}\cmidrule(lr){10-11}\cmidrule(lr){12-13}
    & \textbf{Cle} & \textbf{Ran.}
    & \textbf{Cle} & \textbf{Ran.}
    & \textbf{Cle} & \textbf{Ran.}
    & \textbf{Cle} & \textbf{Ran.}
    & \textbf{Cle} & \textbf{Ran.}
    & \textbf{Cle} & \textbf{Ran.} \\
    \midrule
    \endfirsthead
    \caption[]{Per-task success rates (\%) on RoboTwin 2.0 under the Clean and Random settings
    (continued).} \\
    \toprule
    \multirow{2}{*}{\textbf{Task}}
    & \multicolumn{2}{c}{$\boldsymbol{\pi_0}$}
    & \multicolumn{2}{c}{$\boldsymbol{\pi_{0.5}}$}
    & \multicolumn{2}{c}{\textbf{X-VLA}}
    & \multicolumn{2}{c}{\textbf{Motus}}
    & \multicolumn{2}{c}{\textbf{ZR-0}}
    & \multicolumn{2}{c}{\textbf{\VLADelta{}}} \\
    \cmidrule(lr){2-3}\cmidrule(lr){4-5}\cmidrule(lr){6-7}
    \cmidrule(lr){8-9}\cmidrule(lr){10-11}\cmidrule(lr){12-13}
    & \textbf{Cle} & \textbf{Ran.}
    & \textbf{Cle} & \textbf{Ran.}
    & \textbf{Cle} & \textbf{Ran.}
    & \textbf{Cle} & \textbf{Ran.}
    & \textbf{Cle} & \textbf{Ran.}
    & \textbf{Cle} & \textbf{Ran.} \\
    \midrule
    \endhead
    \midrule
    \multicolumn{13}{r}{\emph{Continued on the next page}} \\
    \endfoot
    \bottomrule
    \endlastfoot
    AdjustBottle & 99 & 95 & 100 & 99 & 100 & 99 & 89 & 93 & 100 & 99 & 100 & 98 \\
    BeatBlockHammer & 79 & 84 & 96 & 93 & 92 & 88 & 95 & 88 & 85 & 88 & 90 & 86 \\
    BlocksRankingRGB & 80 & 63 & 92 & 85 & 83 & 83 & 99 & 97 & 92 & 91 & 95 & 94 \\
    BlocksRankingSize & 14 & 5 & 49 & 26 & 67 & 74 & 75 & 63 & 70 & 81 & 80 & 74 \\
    ClickAlarmclock & 77 & 68 & 98 & 89 & 99 & 99 & 100 & 100 & 96 & 82 & 92 & 94 \\
    ClickBell & 71 & 48 & 99 & 66 & 100 & 100 & 100 & 100 & 90 & 83 & 100 & 100 \\
    DumpBinBigbin & 88 & 83 & 92 & 97 & 79 & 77 & 95 & 91 & 94 & 93 & 97 & 96 \\
    GrabRoller & 98 & 94 & 100 & 100 & 100 & 100 & 100 & 100 & 100 & 100 & 98 & 94 \\
    HandoverBlock & 47 & 31 & 66 & 57 & 73 & 37 & 86 & 73 & 93 & 87 & 93 & 92 \\
    HandoverMic & 97 & 97 & 98 & 97 & 0 & 0 & 78 & 63 & 100 & 99 & 95 & 98 \\
    HangingMug & 14 & 11 & 18 & 17 & 23 & 27 & 38 & 38 & 35 & 33 & 40 & 34 \\
    LiftPot & 80 & 72 & 96 & 85 & 99 & 100 & 96 & 99 & 96 & 98 & 96 & 100 \\
    MoveCanPot & 68 & 48 & 51 & 55 & 89 & 86 & 34 & 74 & 85 & 81 & 94 & 78 \\
    MovePillbottlePad & 67 & 46 & 84 & 61 & 73 & 71 & 93 & 96 & 98 & 99 & 97 & 100 \\
    MovePlayingcardAway & 74 & 65 & 96 & 84 & 93 & 98 & 100 & 96 & 99 & 94 & 96 & 98 \\
    MoveStaplerPad & 41 & 24 & 56 & 42 & 78 & 73 & 83 & 85 & 85 & 92 & 87 & 91 \\
    OpenLaptop & 71 & 81 & 90 & 96 & 93 & 100 & 95 & 91 & 96 & 99 & 98 & 90 \\
    OpenMicrowave & 4 & 32 & 34 & 77 & 79 & 71 & 95 & 91 & 94 & 92 & 95 & 94 \\
    PickDiverseBottles & 69 & 31 & 81 & 71 & 58 & 36 & 90 & 91 & 90 & 88 & 93 & 92 \\
    PickDualBottles & 59 & 37 & 93 & 63 & 47 & 36 & 96 & 90 & 97 & 98 & 92 & 92 \\
    PlaceA2BLeft & 43 & 47 & 87 & 82 & 48 & 49 & 82 & 79 & 83 & 80 & 85 & 84 \\
    PlaceA2BRight & 39 & 34 & 87 & 84 & 36 & 36 & 90 & 87 & 86 & 87 & 85 & 89 \\
    PlaceBreadBasket & 62 & 46 & 77 & 64 & 81 & 71 & 91 & 94 & 90 & 93 & 91 & 95 \\
    PlaceBreadSkillet & 66 & 49 & 85 & 66 & 77 & 67 & 86 & 83 & 92 & 85 & 90 & 89 \\
    PlaceBurgerFries & 81 & 76 & 94 & 87 & 94 & 94 & 98 & 98 & 98 & 98 & 96 & 100 \\
    PlaceCanBasket & 55 & 46 & 62 & 62 & 49 & 52 & 81 & 76 & 66 & 62 & 80 & 64 \\
    PlaceCansPlasticbox & 63 & 45 & 94 & 84 & 97 & 98 & 98 & 94 & 86 & 85 & 94 & 96 \\
    PlaceContainerPlate & 97 & 92 & 99 & 95 & 97 & 95 & 98 & 99 & 99 & 98 & 97 & 100 \\
    PlaceDualShoes & 59 & 51 & 75 & 75 & 79 & 88 & 93 & 87 & 90 & 95 & 92 & 96 \\
    PlaceEmptyCup & 91 & 85 & 100 & 99 & 100 & 98 & 99 & 98 & 97 & 97 & 96 & 99 \\
    PlaceFan & 66 & 71 & 87 & 85 & 80 & 75 & 91 & 87 & 85 & 78 & 84 & 83 \\
    PlaceMousePad & 20 & 20 & 60 & 39 & 70 & 70 & 66 & 68 & 85 & 83 & 87 & 86 \\
    PlaceObjectBasket & 67 & 70 & 80 & 76 & 44 & 39 & 81 & 87 & 75 & 77 & 75 & 79 \\
    PlaceObjectScale & 57 & 52 & 86 & 80 & 52 & 74 & 88 & 85 & 88 & 89 & 89 & 93 \\
    PlaceObjectStand & 82 & 68 & 91 & 85 & 86 & 88 & 98 & 97 & 90 & 91 & 90 & 88 \\
    PlacePhoneStand & 49 & 53 & 81 & 81 & 88 & 87 & 87 & 86 & 85 & 81 & 84 & 83 \\
    PlaceShoe & 76 & 76 & 92 & 93 & 96 & 95 & 99 & 97 & 98 & 97 & 97 & 100 \\
    PressStapler & 44 & 37 & 87 & 83 & 92 & 98 & 93 & 98 & 90 & 92 & 92 & 90 \\
    PutBottlesDustbin & 65 & 56 & 84 & 79 & 74 & 77 & 81 & 79 & 82 & 79 & 82 & 81 \\
    PutObjectCabinet & 73 & 60 & 80 & 79 & 46 & 48 & 88 & 71 & 82 & 76 & 81 & 80 \\
    RotateQRCode & 74 & 70 & 89 & 87 & 34 & 33 & 89 & 73 & 78 & 85 & 83 & 87 \\
    ScanObject & 55 & 42 & 72 & 65 & 14 & 36 & 67 & 66 & 86 & 85 & 88 & 87 \\
    ShakeBottleHorizontally & 98 & 92 & 99 & 99 & 100 & 100 & 100 & 98 & 100 & 100 & 100 & 98 \\
    ShakeBottle & 94 & 91 & 99 & 97 & 99 & 100 & 100 & 97 & 100 & 100 & 100 & 100 \\
    StackBlocksThree & 72 & 52 & 91 & 76 & 6 & 10 & 91 & 95 & 86 & 88 & 88 & 92 \\
    StackBlocksTwo & 93 & 79 & 97 & 100 & 92 & 87 & 100 & 98 & 95 & 91 & 94 & 93 \\
    StackBowlsThree & 77 & 75 & 77 & 71 & 76 & 86 & 79 & 87 & 79 & 88 & 83 & 87 \\
    StackBowlsTwo & 94 & 95 & 95 & 96 & 96 & 93 & 98 & 98 & 94 & 92 & 97 & 96 \\
    StampSeal & 46 & 33 & 79 & 55 & 76 & 82 & 93 & 92 & 92 & 92 & 92 & 95 \\
    TurnSwitch & 41 & 42 & 62 & 54 & 40 & 61 & 84 & 78 & 83 & 78 & 83 & 82 \\
    \midrule
    \textbf{Average} & \textbf{65.92} & \textbf{58.4} & \textbf{82.74} & \textbf{76.76} & \textbf{72.88} & \textbf{72.84} & \textbf{88.52} & \textbf{87.02} & \textbf{88.7} & \textbf{87.98} & \textbf{90.06} & \textbf{89.74} \\
\end{longtable}
\endgroup

%% file: tables/table_real_robot_tasks.tex
\begin{table}[htbp]
    \centering
    \caption{Capability dimensions and task instructions for real-robot evaluation.}
    \label{tab:real_robot_tasks}
    \small
    \begin{tabularx}{\linewidth}{@{}>{\raggedright\arraybackslash}p{0.40\linewidth}>{\raggedright\arraybackslash}X@{}}
        \toprule
        \multicolumn{1}{c}{\textbf{Capability Dimension}}
        & \multicolumn{1}{c}{\textbf{Task Instruction}} \\
        \midrule
        Obstacle Avoidance and Physical Constraint Compliance
        & Place the carrots on a rack, making sure they do not touch the cups. \\
        \addlinespace
        Long-Horizon Execution
        & Put all the items on the tabletop onto the tray. \\
        \addlinespace
        Fine-Grained Manipulation
        & Put the spoon into the cup. \\
        \addlinespace
        Spatial Reasoning
        & Place the item closest to the bowl onto the plate. \\
        \addlinespace
        Generalization under Changed Tablecloth
        & Place the item closest to the bowl onto the plate. \\
        \bottomrule
    \end{tabularx}
\end{table}

%% file: sections/appendix_training.tex
\section{Training Details}
\label{sec:appendix-training}

\subsection{Structured Slot Heads and Supervision Objectives}
\label{sec:slot-head-details}

\textbf{Prediction head architecture and outputs.}
Each prediction branch consists of LayerNorm followed by a single linear projection:
\begin{equation}
u_{t,i}
= W_i\,\operatorname{LN}_i\!\left(\bar z_{t,g(i)}\right)+b_i,
\label{eq:slot-head-projection}
\end{equation}
where $g(i)$ identifies the Query group assigned to target $S_i$, and
$\bar z_{t,g(i)}$ is the mean-pooled final Query representation of that group. Depending on the
target type, the output is then reshaped, passed through a sigmoid function, or transformed using
a valid bounding-box parameterization.

$S_1$ and $S_2$ share the pooled representation but have separate LayerNorm and linear layers.
$S_7$ and $S_8$ share both the pooled representation and LayerNorm, i.e.,
$\operatorname{LN}_7=\operatorname{LN}_8$, while using two independent linear layers. The remaining
branches each have their own normalization and linear projection parameters. The complete module
contains nine linear layers and eight LayerNorm layers, with no additional activation function or
dropout between normalization and linear projection.

Task progress ($S_1$) and task phase ($S_2$) describe task progression through continuous completion
levels and discrete manipulation stages, respectively, and therefore share the Query
representation of G12. Gripper state transitions ($S_7$) and contact transitions ($S_8$) both
describe interaction events during manipulation and therefore share the Query representation of
G78. By assigning semantically related supervision targets to the same Query group, we encourage
the model to jointly encode task progression and interaction changes while retaining
target-specific prediction capabilities through separate output branches. This sharing provides
a soft semantic division of labor without requiring a deterministic one-to-one correspondence
between related targets.

Table~\ref{tab:structured_slot_outputs} summarizes the outputs of the nine target categories.
All output shapes exclude the batch dimension.

\input{tables/table_structured_slot_outputs}

Because Queries within each group are mean-pooled, all output components are jointly predicted
from the same group representation. For example, the target-position branch predicts bounding
boxes at two time steps, but no Query is exclusively assigned to the current box or the future
box. Contact points, three-dimensional displacements, and obstacle positions follow the same
readout scheme.

Task phase ($S_2$), gripper state transitions ($S_7$), and contact transitions ($S_8$) use fixed
class vocabularies. Table~\ref{tab:structured_slot_vocabularies} lists the classes in the order of
their zero-based class IDs.

\input{tables/table_structured_slot_vocabularies}

%% file: tables/table_structured_slot_outputs.tex
\begingroup
\small
\setlength{\tabcolsep}{3pt}
\setlength{\LTcapwidth}{\textwidth}
\renewcommand{\arraystretch}{1.0}
\begin{longtable}{@{}>{\raggedright\arraybackslash}p{0.18\linewidth}>{\centering\arraybackslash}p{0.08\linewidth}>{\raggedright\arraybackslash}p{0.23\linewidth}>{\centering\arraybackslash}p{0.10\linewidth}>{\raggedright\arraybackslash}p{\dimexpr0.41\linewidth-8\tabcolsep\relax}@{}}
    \caption{Structured Transition Slot outputs, Query groups, and supervision. Output shapes exclude the batch dimension.}
    \label{tab:structured_slot_outputs} \\
    \toprule
    \textbf{Supervision Target} & \textbf{Query Group} & \textbf{Output Content}
    & \textbf{Output Shape} & \textbf{Output Constraints and Supervision} \\
    \midrule
    \endfirsthead
    \caption[]{Structured Transition Slot outputs, Query groups, and supervision (continued).} \\
    \toprule
    \textbf{Supervision Target} & \textbf{Query Group} & \textbf{Output Content}
    & \textbf{Output Shape} & \textbf{Output Constraints and Supervision} \\
    \midrule
    \endhead
    \midrule
    \multicolumn{5}{r}{\emph{Continued on the next page.}} \\
    \endfoot
    \bottomrule
    \endlastfoot
    $S_1$: Task progress & G12
    & Task progress at the current and next keyframes & $2$
    & Sigmoid; Smooth L1 and a progress monotonicity penalty \\
    \addlinespace[2pt]
    $S_2$: Task phase & G12
    & Seven manipulation phases at two time steps & $2\times7$
    & Classification logits; class-weighted cross-entropy with label smoothing \\
    \addlinespace[2pt]
    $S_3$: Target position & G3
    & Target-object boxes at two time steps & $2\times4$
    & Valid box parameterization; Smooth L1 and GIoU \\
    \addlinespace[2pt]
    $S_4$: Gripper position & G4
    & Gripper boxes at two time steps & $2\times4$
    & Valid box parameterization; Smooth L1 and GIoU \\
    \addlinespace[2pt]
    $S_5$: Motion and relative displacement & G5
    & Target, gripper, and relative 3D displacements & $3\times3$
    & Standardized regression and displacement consistency in meters \\
    \addlinespace[2pt]
    $S_6$: Interactable position & G6
    & Two affordance contact points in the current frame & $2\times2$
    & Sigmoid; pointwise Smooth L1 \\
    \addlinespace[2pt]
    $S_7$: Gripper state transitions & G78
    & Four gripper opening/closing transition classes & $4$
    & Classification logits; class-weighted cross-entropy with label smoothing \\
    \addlinespace[2pt]
    $S_8$: Contact transitions & G78
    & Four contact-transition classes & $4$
    & Classification logits; class-weighted cross-entropy with label smoothing \\
    \addlinespace[2pt]
    $S_9$: Obstacles and risk & G9
    & Boxes, risk scores, and validity for three ordered annotation positions & $3\times6$
    & Box loss, risk regression, and binary validity classification \\
\end{longtable}
\endgroup

%% file: tables/table_structured_slot_vocabularies.tex
\begingroup
\small
\setlength{\tabcolsep}{3pt}
\setlength{\LTcapwidth}{\textwidth}
\renewcommand{\arraystretch}{1.12}
\begin{longtable}{@{}>{\raggedright\arraybackslash}p{0.25\linewidth}>{\raggedright\arraybackslash}p{\dimexpr0.75\linewidth-2\tabcolsep\relax}@{}}
    \caption{Fixed class vocabularies for task phase, gripper state transitions, and contact transitions. Class IDs start at zero.}
    \label{tab:structured_slot_vocabularies} \\
    \toprule
    \textbf{Target} & \textbf{Classes in ID Order and Meanings} \\
    \midrule
    \endfirsthead
    \caption[]{Fixed class vocabularies for task phase, gripper state transitions, and contact transitions (continued).} \\
    \toprule
    \textbf{Target} & \textbf{Classes in ID Order and Meanings} \\
    \midrule
    \endhead
    \midrule
    \multicolumn{2}{r}{\emph{Continued on the next page.}} \\
    \endfoot
    \bottomrule
    \endlastfoot
    $S_2$: Task phase
    & 0: \texttt{align} (alignment);
      1: \texttt{approach} (approaching);
      2: \texttt{contact} (contact);
      3: \texttt{manipulate} (object manipulation);
      4: \texttt{release} (release);
      5: \texttt{retract} (retraction);
      6: \texttt{transport} (transport) \\
    \addlinespace
    $S_7$: Gripper state transitions
    & 0: \texttt{closed\_to\_open} (closed to open);
      1: \texttt{maintain\_closed} (remaining closed);
      2: \texttt{maintain\_open} (remaining open);
      3: \texttt{open\_to\_closed} (open to closed) \\
    \addlinespace
    $S_8$: Contact transitions
    & 0: \texttt{break\_contact} (breaking contact);
      1: \texttt{establish\_contact} (establishing contact);
      2: \texttt{maintain\_contact} (maintaining contact);
      3: \texttt{remain\_separated} (remaining separated) \\
\end{longtable}
\endgroup